\documentclass[11pt, a4paper, logo,  copyright]{googledeepmind}

\usepackage[authoryear, sort&compress, round]{natbib}
\usepackage{listings}
\usepackage{tcolorbox}
\usepackage{caption}
\usepackage{booktabs}

\title{Evaluation of retrieval-based QA on QUEST-LOFT}

\renewcommand{\today}{2025-11-04}

\author[*,1]{Nathan Scales}
\author[*,1]{Nathanael Sch\"arli}
\author[1]{Olivier Bousquet}

\affil[*]{Equal contributions}
\affil[1]{Google DeepMind}

\begin{abstract}
Despite the popularity of retrieval-augmented generation (RAG) as a solution for grounded QA in both academia and industry, current RAG methods struggle with questions where the necessary information is distributed across many documents or where retrieval needs to be combined with complex reasoning. Recently, the LOFT study has shown that this limitation also applies to approaches based on long-context language models, with the QUEST benchmark exhibiting particularly large headroom. In this paper, we provide an in-depth analysis of the factors contributing to the poor performance on QUEST-LOFT, publish updated numbers based on a thorough human evaluation, and demonstrate that RAG can be optimized to significantly outperform long-context approaches when combined with a structured output format containing reasoning and evidence, optionally followed by answer re-verification.
\end{abstract}

\begin{document}

\maketitle

\section{Introduction}
\label{sect:introduction}

Large language models (LLMs) are known to be able to internalize into their weights significant amounts of information seen in training. However, LLMs' parametric knowledge alone is insufficient for use cases involving long-tail entities, fresh information, and information stored in private or specialized corpora, and more generally for use cases for which the risk of hallucination is unacceptable~\citep{min2023factscore, huang2023survey}. In such cases, the commonly-used approach currently is retrieval-augmented generation (RAG)~\citep{lewis2020rag, gao2023rag}, in which relevant content is retrieved from a corpus, either all at once or iteratively, and then provided to the LLM as context. Using this context, the LLM is typically prompted to summarize or answer questions based on the retrieved content. This approach has, on the one hand, led to strong results on QA tasks in which the answers can be synthesized from a small number of documents in a relatively straightforward fashion. However, RAG struggles on questions where the necessary information is distributed across many documents or where retrieval needs to be combined with complex reasoning.

More recently, the advent of LLMs with context windows of 1M tokens or more has opened up new possibilities for dealing with factuality, in which the LLM can be provided in its context a much larger quantity of retrieved information (\textit{long-context}), or possibly even an entire corpus (\textit{corpus-in-context}). In studies such as the one presented with the Long-Context Frontiers (LOFT) benchmark~\citep{lee2024loft}, however, it has been shown that corpus-in-context solutions suffer from many of the same limitations mentioned above for RAG. Specifically, they show that performance on most datasets is roughly comparable on corpora of 128K tokens, but that corpus-in-context underperforms RAG noticeably at the 1M token scale. In this paper, we dive deeper into factors contributing to this headroom and possible improved solutions, with a focus on the QA dataset that showed the highest headroom in the LOFT study, namely the QUEST benchmark~\citep{malaviya2023quest}. For this dataset, they report subspan exact match accuracy of only 0.28 for corpus-in-context QA and only 0.35 for a RAG solution on the 128K token version of the dataset using Gemini 1.5 Pro.

Specifically, we make the following contributions:
\begin{itemize}
    \item Provide a revised and expanded set of ground truth answers for QUEST-LOFT based on comprehensive human evaluation.
    \item Evaluate a range of techniques on the revised QUEST-LOFT dataset, including structured outputs, chain-of-thought, and self-verification, in addition to the corpus-in-context and retrieve-and-read baselines.
    \item Improve SOTA on QUEST-LOFT-128K-Revised (F1 +0.16, accuracy +0.16, subspan EM +0.11) through a combination of RAG with structured outputs and self-verification.
    \item Reinforce the findings of the LOFT paper regarding the out-performance of retrieval-based solutions compared to corpus-in-context, while demonstrating an even larger performance gap.
    \item Reinforce the findings of prior research that most -- though not always all -- of the benefits of self-verification can be equivalently achieved via improvements to the prompt used in generating the initial response.
    \item Analyze remaining headroom and suggest roadmap for future work.
\end{itemize}

\section{Datasets}
\label{sect:dataset}

In this paper, we report results on evaluations of two new variations of the QUEST-LOFT-128K dataset (part of the LOFT benchmark suite), which we call \textit{QUEST-LOFT-128K-Revised} and \textit{QUEST-LOFT-128K-Simple28}. For context, we first introduce here the original QUEST~\citep{malaviya2023quest} dataset and the LOFT~\citep{lee2024loft} benchmark suite, followed by our newly-introduced variations.

For a summary of the dataset stats, see Table~\ref{tab:datasets}.

\begin{table}[htbp]
\footnotesize
\centering
\begin{tabular}{@{}l|rrrl@{}}
\toprule
Dataset & Test & Dev & Entities & Notes \\
\midrule
QUEST & 1,307 & 323 & 325,505 & Introduced by \cite{malaviya2023quest}.\\
QUEST-LOFT-128K & 100 & 10 & 328 & Introduced by \cite{lee2024loft}. \\
\midrule
QUEST-LOFT-128K-Revised & 100 & 10 & 328 & Same questions and entities as QUEST-LOFT-128K. \\
QUEST-LOFT-128K-Simple28 & 28 & -- & 328 & Same entities as QUEST-LOFT-128K. \\
\bottomrule
\end{tabular}
\caption{Overview of datasets. The top group of datasets are the existing datasets, described here for context. The bottom group are the newly-introduced variations for which we report results.}
\label{tab:datasets}
\end{table}

\subsection{QUEST}
\label{sect:quest}

The original QUEST dataset consists of a corpus of 325,505 Wikipedia pages (describing 325,505 entities), paired with a set of 3,357 questions (1,307 train, 323 validation, 1,727 test), whose answers take the form of a set of up to 20 entities from that corpus. The question-answer pairs were automatically generated based on the Wikipedia category names associated with each Wikipedia document, together with the application of one of 7 implicit set operation templates, and then vetted by human raters with the intention of ensuring that the questions are indeed answerable from the document content alone. The documents in the corpus are represented as plain text (without images or markup, and without additional metadata such as the Wikipedia category names that were used in the generation of the questions). Question metadata is provided for debugging purposes, including evidence provided by the human raters supporting the answers. (For an example, see Figure~\ref{fig:quest_example}.)

\begin{figure}[ht]
\begin{tcolorbox}[colback=gray!5!white,colframe=gray!25!black,arc=0pt,outer arc=0pt,boxrule=0.4pt,toprule=0.4pt,bottomrule=0.4pt,leftrule=0.4pt,rightrule=0.4pt]
  \small\textbf{Question:} Action comedy films from Taiwan or comedy films about martial arts from China \\
  \small\textbf{Answer:} 
  \texttt{[`The Adventures of Jinbao', `The Invincible Constable', `The Village of No Return']}
\end{tcolorbox}
\captionsetup{font=footnotesize}
\caption{\footnotesize QUEST dataset example}
\label{fig:quest_example}
\end{figure}

\subsection{LOFT}
\label{sect:loft}

The LOFT benchmark suite consists of modified versions of 36 pre-existing datasets across 6 tasks, including 6 datasets for the ``RAG'' (i.e., corpus-based QA) task. Each dataset is filtered down to 100 test questions, 10 dev questions, and 5 train questions (for use as few-shot exemplars), together with a paired-down corpus that comes in 3 sizes designed to fit inside of a 32K, 128K, and 1M token context window respectively. Of these, the 128K token corpus is the one on which the LOFT paper presents its primary results and is also the one that we focus on in this paper. For the QUEST dataset, the 128K token corpus consists of relevant snippets from the Wikipedia pages corresponding to 328 candidate entities. We will refer to the versions of QUEST provided with the LOFT benchmark collectively as QUEST-LOFT and the variant with the 128K token corpus as QUEST-LOFT-128K.

One thing to note is that, while the questions appearing in QUEST-LOFT-128K are a strict subset of the questions from the original QUEST dataset, performance numbers on these datasets are not directly comparable, due to the following differences:

\begin{itemize}
	\item The retrieval task is significantly easier on QUEST-LOFT-128K than on QUEST due to the several orders of magnitude difference in the size of the corpus (e.g., 325,505 entities for QUEST vs. 328 entities for QUEST-LOFT-128K).
	\item Even for the entities that are in scope for QUEST-LOFT-128K, the QUEST-LOFT-128K corpus contains less information than is available in the full QUEST corpus, as the construction of QUEST-LOFT involved splitting the Wikipedia page for each entity into passages of a roughly uniform size and then keeping only the ``golden'' passages, i.e., the passages that had been indicated in the QUEST metadata as being relevant to answering the given question. The effect of this passage filtering on task difficulty can be subtle. On the one hand, it can conceivably make the task easier, as it further reduces the amount of distracting information that the model needs to ignore. However, we also observed in manual analysis of several of the questions in the dataset that this filtering sometimes led to removal of content that could potentially be relevant to the answering of the question, or made the content of the remaining passages more difficult to interpret out of context.
	\item In QUEST-LOFT-128K, each of the golden passages is treated as a separate document, which means that unlike in the original QUEST dataset where there is exactly one document in the corpus for each entity, in QUEST-LOFT-128K there can be multiple. In our experiments, we observed this to empirically make the task more difficult for the LLM in both the long context and RAG setups. (See \ref{sect:dataset_representation} for mitigations.)
\end{itemize}

\subsection{New variations of QUEST-LOFT-128K}
\label{sect:quest_loft_128k_new_variations}

The main results of this paper are based on a variation of QUEST-LOFT-128K, which we call \textit{QUEST-LOFT-128K-Revised}. This dataset contains the same questions and documents as QUEST-LOFT-128K, but with revised golden answers that we prepared based on a comprehensive human evaluation. We describe the process used for golden answer curation, along with experiment results in Section~\ref{sect:quest_loft_128k_revised}.

To better explore the degree to which the failures on QUEST-LOFT-128K-Revised are due to the ability to understand and reason about complex questions vs.\ the ability to accurately determine the underlying atomic facts, we also manually constructed a small dataset consisting of 28 atomic questions (i.e., ones that do not require set operations), based on a decomposition of the 10 questions from the QUEST-LOFT-128K dev set, while continuing to use the same corpus. We refer to this dataset as \textit{QUEST-LOFT-128K-Simple28} and present the curation process and experiment results in Section~\ref{sect:quest_loft_128k_simple28}. 
  
\section{Experimental Setup}
\label{sect:experimental_setup}

\subsection{Dataset Representation}
\label{sect:dataset_representation}

As a rule, in our experiments, we provide the inputs for each example (i.e., question + documents) to each evaluated solution in the same format as in the LOFT paper, with the exception that instead of representing each passage as a separate ``document'', we merge together all of the passages that originated from the same Wikipedia page, in the same order in which they originally appeared in the page, so as to ensure that the task is not made unnecessarily difficult due to passages appearing out of context. The merged representation is also convenient, in that it ensures that there is exactly one document per candidate entity (the same as in the original QUEST dataset) and that we can hence use the document ID equivalently as a unique identifier of the corresponding entity (i.e., as an ``entity ID'').

\subsection{Evaluation Framework}
\label{sect:evaluation_framework}

To ensure careful control and a fair comparison of the solutions that we evaluate, we implement all solutions in a single integrated Python codebase that we build using the library OneTwo~\citep{onetwo2024github}. This includes re-implementations of the \textit{retrieve-and-read} and \textit{corpus-in-context} QA solutions from the LOFT paper, as well as new variants that we introduce below. In order to facilitate ablation studies and recombination of components, we found it useful to factor solutions into the following types of components, which can be freely mixed and matched:

\begin{itemize}
    \item \textbf{Indexing+retrieval strategy}: Takes as input a corpus in the form of a sequence of documents and constructs from that a \textit{retriever} that can retrieve the top K most relevant documents for a given query (specified as a \textit{max\_results} parameter). In the simplest case, the documents can simply be stored as-is and the same K documents returned regardless of the query, or if \textit{max\_results} is omitted, then the retriever will return the entire contents of the corpus. The retriever only needs to be constructed once for the dataset as a whole and can then be shared across all questions.
    \item \textbf{Retrieval QA strategy}: Takes as input a question and a retriever and returns an answer in the form of a string or list of strings, or a more complex data structure from which this can be extracted. (In the case of QUEST, the answer is represented as a list of entity names, which are equivalent to the titles of the corresponding Wikipedia documents.)
    \item \textbf{Answer verification strategy}: Takes as input a question and a candidate answer, along with a limited number of directly relevant documents. Returns a judgment of whether the candidate answer is a valid answer to the question.
\end{itemize}

In our experiments, we perform evaluation sweeps over different combinations and configurations of these components.

For an overview of the methods evaluated, see Table~\ref{tab:evaluated_methods}.

\begin{table}[htbp]
\footnotesize
\centering
\begin{tabular}{@{}l|lll@{}}
\toprule
Method & Indexing+retrieval & QA & Verification \\
\midrule
CiC Baseline & Static - All & CiC & -- \\
CiC Justified QA & Static - All & Justified & -- \\
CiC Justified QA + CoT & Static - All & Justified - CoT & -- \\
CiC Justified QA + Verification & Static - All & Justified & Basic \\
CiC Justified QA + CoT + Verification & Static - All & Justified - CoT & CoT \\
CiC Justified QA + Verification + QUEST & Static - All & Justified - QUEST & QUEST \\
CiC Justified QA + CoT + Verification + QUEST & Static - All & Justified - CoT - QUEST & CoT - QUEST \\
\midrule
RAG Baseline & Embedding - Top 40 & RaR & -- \\
RAG Justified QA & Embedding - Top 40 & Justified & -- \\
RAG Justified QA + CoT & Embedding - Top 40 & Justified - CoT & -- \\
RAG Justified QA + Verification & Embedding - Top 40 & Justified & Basic \\
RAG Justified QA + CoT + Verification & Embedding - Top 40 & Justified - CoT & CoT \\
RAG Justified QA + Verification + QUEST & Embedding - Top 40 & Justified - QUEST & QUEST \\
RAG Justified QA + CoT + Verification + QUEST & Embedding - Top 40 & Justified - CoT - QUEST & CoT - QUEST \\
\midrule
RAG + Verification & Embedding - Top 40 & -- & Basic \\
RAG + Verification (w/ CoT) & Embedding - Top 40 & -- & CoT \\
RAG + Verification + QUEST & Embedding - Top 40 & -- & QUEST \\
RAG + Verification (w/ CoT) + QUEST & Embedding - Top 40 & -- & CoT - QUEST \\
\bottomrule
\end{tabular}
\caption{Methods evaluated in main experiments.}
\label{tab:evaluated_methods}
\end{table}

\subsection{Models}
\label{sect:models}

All QA strategies were evaluated using Gemini 1.5 Pro~\citep{team2024gemini}, except where otherwise indicated. For RAG strategies, retrieval was performed using brute force dot product similarity on embeddings from Gecko~\citep{lee2024gecko} version  ``text-embedding-004''. These settings match those used in the original LOFT experiments in \cite{lee2024loft}.

\subsection{Indexing+retrieval strategies}
\label{sect:indexing_retrieval_strategies}

Consistent with the approaches taken in \cite{lee2024loft}, we evaluate the following indexing+retrieval strategies:
\begin{itemize}
    \item \textbf{Static retrieval}: Performs no special indexing and always returns the same set of documents. Specifically, we use this approach for corpus-in-context strategies that simply include the entire corpus in the context, regardless of the question.
    \item \textbf{Embedding-based retrieval}: Indexes the documents using embeddings calculated from their title and text content and then retrieves the K nearest neighbors based on dot product similarity between the query embedding and document embeddings.
\end{itemize}

\subsection{Retrieval QA Strategies}
\label{sect:retrieval_qa_strategies}

As baselines, we evaluate the following two strategies described in  \cite{lee2024loft}, which we re-implement with minor modifications:
\begin{itemize}
    \item \textbf{Corpus-in-context (CiC)}: Obtains from the retriever the entire contents of the corpus, by querying it with the \textit{max\_results} constraint omitted. Then prompts the LLM with the corpus contents included in the preamble, followed by 5 few-shot question-answer pairs from the train set, and then finally the actual question of interest.
    \item \textbf{Retrieve-and-read (RaR)}: Obtains from the retriever the top 40 most relevant documents for the question of interest, as well as for each of the 5 few-shot exemplars from the train set. Then prompts the LLM with the 5 few-shot question-context-answer triples, followed by the actual context and the actual question of interest. 
\end{itemize}

In both of the above strategies, the LLM is prompted to output a list of entity IDs (i.e., document IDs), which we then convert to a list of entity names (i.e., document titles) programmatically via post-processing. (See Figure~\ref{fig:baseline_qa_output} for an example of the output format.)

\begin{figure}[ht]
\begin{tcolorbox}[colback=gray!5!white, colframe=gray!25!black, arc=0pt, outer arc=0pt, boxrule=0.4pt, toprule=0.4pt, bottomrule=0.4pt, leftrule=0.4pt, rightrule=0.4pt]
\lstset{basicstyle=\ttfamily\footnotesize, breaklines=true}
\begin{lstlisting}[language=, basicstyle=\ttfamily\footnotesize, breaklines=true]
The following documents are needed to answer the query:
TITLE: The Adventures of Jinbao | ID: 192
TITLE: The Invincible Constable | ID: 74
TITLE: The Village of No Return | ID: 77
Final Answer: ['192', '74', '77']
\end{lstlisting}
\end{tcolorbox}
\captionsetup{font=footnotesize}
\caption{\footnotesize Example of LLM output for the baseline QA strategies.}
\label{fig:baseline_qa_output}
\end{figure}
                    
One limitation of the above baselines is that when their outputs differ from the golden answers, it is non-trivial to analyze the root cause for the discrepancy, as the output provides no transparency into how the LLM arrived at the given final answer. To address this limitation, we further evaluate the following \textit{Justified QA} strategies, which prompt the LLM to output a JSON data structure containing the final answers together with detailed provenance information.

\begin{itemize}
    \item \textbf{Justified}: Prompts the LLM with  instructions on how to answer the question with care, while outputting its evidence and reasoning steps in the form of a JSON data structure. The prompt contains the detailed instructions, following by the retrieved documents and the question of interest. In lieu of few-shot exemplars, we simply provide an example of the output format. By default, we use a generic instruction that could be applied to an arbitrary short-form or list-answered QA task.
    \item \textbf{Justified - CoT}: Variant of Justified QA, in which the LLM is prompted to first output its reasoning in natural language form (i.e., as a traditional \textit{chain-of-thought}), prior to outputting the JSON data structure.
    \item \textbf{Justified - QUEST}: Variant of Justified QA, in which the prompt contains QUEST-specific instructions tailored to the somewhat idiosyncratic format of the questions and answers in QUEST.
    \item \textbf{Justified - CoT - QUEST}: Variant of Justified QA - CoT, in which the prompt includes the same QUEST-specific instructions as in Justified QA - QUEST.
\end{itemize}

To avoid the need for fuzzy matching between entity names during metric calculation, we apply the same technique used in the CiC and RaR QA strategies above, in which we have the LLM output a final answer in the form of a list of entity IDs (i.e., document IDs), which we then convert to a list of entity names (i.e., document titles) programmatically.

See Figure~\ref{fig:verifiable_qa_output} for an example of the format of the JSON data structure output by the above strategies. For details of the prompts used, see Appendix~\ref{asect:verifiable_qa}.

\begin{figure}[ht]
\begin{tcolorbox}[colback=gray!5!white, colframe=gray!25!black, arc=0pt, outer arc=0pt, boxrule=0.4pt, toprule=0.4pt, bottomrule=0.4pt, leftrule=0.4pt, rightrule=0.4pt]
\lstset{basicstyle=\ttfamily\footnotesize, breaklines=true}
\begin{lstlisting}[language=, basicstyle=\ttfamily\footnotesize, breaklines=true]
{
  "type": "VerifiableContextualQAResponse",
  "question": "1990s Indian and folklore romance films.",
  "candidate_answers": [
    {
      "candidate_answer": "Roja (film)",
      "evidence_for": [
        {
          "doc_id": "75",
          "text": "'''''Roja''''' () is a 1992 Indian Tamil-language romantic thriller film written and directed by Mani Ratnam.... It follows a simple girl from a village in Tamil Nadu, making desperate efforts to find her husband after he is kidnapped by militants during a secret undercover mission in Jammu and Kashmir."
        }
      ],
      "evidence_against": [],
      "reasoning": "Roja (film) is a 1992 Indian romance film, thus satisfying the criteria of being an Indian romance film from the 1990s.",
      "final_judgment": "TRUE"
    },
    ...
  ],
  "answer": ["Roja (film)", "Sahasa Veerudu Sagara Kanya"],
  "answer_doc_ids": ["75", "220"]
}
\end{lstlisting}
\end{tcolorbox}
\caption{\footnotesize Example of LLM output for a Justified QA strategy.}
\label{fig:verifiable_qa_output}
\end{figure}

\subsection{Answer Verification Strategies}
\label{sect:answer_verification_strategies}

Taking inspiration from the Chain-of-Verification (CoVe) strategy of \cite{dhuliawala2023cove}, we further apply an optional answer verification step, in which we gather all of the candidate answers that were output by the original Justified QA step (including ones for which the Justified QA step had initially indicated a final judgment of `FALSE') and then prompt the LLM to make an independent judgment for each of the candidate answers individually as to whether it is a valid answer to the given question, based purely on the document(s) that were cited by the Justified QA strategy as evidence for that candidate answer. We evaluate the following answer verification variants:

\begin{itemize}
    \item \textbf{Basic}: By default, we use a generic instruction that could be applied to an arbitrary short-form or list-answered QA task.
    \item \textbf{CoT}: Variant in which the LLM is prompted to first output its reasoning in natural language form (i.e., as a traditional \textit{chain-of-thought}), prior to outputting the JSON data structure.
    \item \textbf{QUEST}: Variant that includes QUEST-specific instructions tailored to the format of the questions and answers in QUEST.
    \item \textbf{CoT - QUEST}: Variant incorporating both chain-of-thought and the QUEST-specific instructions.
\end{itemize}

For details of the prompts used, see Appendix~\ref{asect:answer_verification}.

\subsection{Metrics}
\label{sect:metrics}

Consistent with the nature of QUEST as a multi-value QA task, in which the individual answers can be identified unambiguously via unique entity IDs (i.e., document IDs) or canonical entity names (i.e., document titles), we focus in our evaluations primarily on the standard value-set-based metrics of \textit{F1 score}, \textit{precision}, \textit{recall}, and \textit{accuracy}. For each of these, we calculate the metric first at the level of individual examples, by comparing the set of golden values with the set of values predicted by the QA strategy. We then average the value of the metric across all examples in the dataset to yield the values at the dataset level.

To enable comparison with previously-published numbers, we also calculate \textit{subspan exact match (subspan EM)}, which was the primary metric reported for RAG (i.e., corpus-based QA) tasks in the LOFT paper. The subspan comparison aspect itself is not relevant in our case, since we always ensure (as described in \ref{sect:retrieval_qa_strategies}) that the final answers output by the QA strategy are always canonical entity names which can be compared to the golden answers via simple string equality. In this situation, the \textit{subspan EM} boils down to a variation of \textit{recall}, in which credit is given only when the predicted answer set includes all of the answers from the golden set.

\section{QUEST-LOFT-128K-Revised}
\label{sect:quest_loft_128k_revised}

\subsection{Golden Answer Curation}
\label{sect:quest_loft_128k_revised_curation}

While iterating on the QUEST-LOFT dev set and later during error analysis on samples of the test set, we observed a significant number of issues in the golden set. For a fraction of supposed recall errors, the golden answer did not seem to be fully justified by evidence from the given Wikipedia page (or from other information that we could obtain from web search) or else was debatable, whereas for a fraction of supposed precision errors, the answer proposed by the LLM appeared to be correct and simply missing from the golden answer set. 
In order to obtain a clearer signal in our evaluations, we invested in a comprehensive human evaluation of the limited set of questions and documents included in QUEST-LOFT-128K, so as to obtain a revised set of manually-curated golden answers, which we refer to as \textit{QUEST-LOFT-128K-Revised}.

We performed the human evaluation ourselves via the following procedure for each question:
\begin{enumerate}
    \item Gather as candidate answers the ground truth answers from the original dataset along with the answers from each candidate solution.
    \item For each candidate answer, manually check snippets included in QUEST-LOFT-128K, the full document from the original QUEST corpus, as well as information from sources such as IMDB via a manual web search. While doing this, take into account the evidence cited by the human raters in the original dataset metadata and by the solutions, where applicable.
    \item Depending on the evidence, classify each answer into the buckets NO\_MATCH (the answer clearly does not match the question), DEBATABLE (there are good arguments for and against considering the answer a match for the question), and MATCH (the answer clearly matches the question).
\end{enumerate}

All evaluation results shown in this paper are based on the revised dataset. Specifically, we computed all our metrics after removing the DEBATABLE answer from both the golden and the candidate sets. See Appendix~\ref{asect:quest_loft_128k_revised} for details.

\subsection{Evaluation on Gemini 1.5 Pro}
\label{sect:quest_loft_128k_revised_evaluation_pro}

For results, see the leaderboard in Table~\ref{tab:leaderboard_quest_loft_128k_revised_test_pro}.

\begin{table}[htbp]
\footnotesize
\centering
\begin{tabular}{@{}l|ccccc@{}}
\toprule
Method & F1 & Precision & Recall & Accuracy & Subspan EM \\
\midrule
CiC Baseline & 0.66 & 0.78 & 0.70 & 0.38 & 0.50 \\
CiC Justified QA & 0.74 & 0.88 & 0.74 & 0.42 & 0.53 \\
CiC Justified QA + CoT & 0.74 & 0.87 & 0.72 & 0.45 & 0.53 \\
CiC Justified QA + Verification & 0.76 & 0.93 & 0.71 & 0.45 & 0.50 \\
CiC Justified QA + CoT + Verification & 0.72 & \textbf{0.95} & 0.68 & 0.47 & 0.50 \\
CiC Justified QA + Verification + QUEST & 0.77 & 0.94 & 0.73 & 0.51 & 0.54 \\
CiC Justified QA + CoT + Verification + QUEST & 0.76 & 0.94 & 0.71 & 0.49 & 0.50 \\
\midrule
RAG Baseline & 0.67 & 0.79 & 0.72 & 0.41 & 0.55 \\
RAG Justified QA & 0.81 & 0.92 & 0.79 & 0.55 & 0.64 \\
RAG Justified QA + CoT & 0.76 & 0.90 & 0.75 & 0.47 & 0.57 \\
RAG Justified QA + Verification & \textbf{0.83} & 0.93 & 0.82 & 0.57 & 0.67 \\
RAG Justified QA + CoT + Verification & 0.74 & \textbf{0.95} & 0.71 & 0.47 & 0.54 \\
RAG Justified QA + Verification + QUEST & \textbf{0.83} & 0.93 & 0.82 & 0.57 & 0.66 \\
RAG Justified QA + CoT + Verification + QUEST & 0.80 & 0.94 & 0.77 & \textbf{0.62} & 0.65 \\
\midrule
RAG + Verification & 0.77 & 0.74 & \textbf{0.91} & 0.43 & 0.79 \\
RAG + Verification (w/ CoT) & 0.79 & 0.86 & 0.84 & 0.49 & 0.71 \\
RAG + Verification + QUEST & 0.78 & 0.76 & 0.90 & 0.47 & \textbf{0.80} \\
RAG + Verification (w/ CoT) + QUEST & 0.80 & 0.82 & 0.86 & 0.54 & 0.73 \\
\bottomrule
\end{tabular}
\caption{Evaluation on QUEST-LOFT-128K-Revised with Gemini 1.5 Pro.}
\label{tab:leaderboard_quest_loft_128k_revised_test_pro}
\end{table}

From these results, we can make the following observations:
\begin{itemize}
    \item Metric values based on the revised golden answers are overall considerably higher than those reported in the original LOFT paper. This suggests that a third or more of the ``errors'' seen in the original experiments were actually cases in which the golden answer was incorrect or debatable.
    \item Similarly to the original LOFT study, we find that the RAG baseline outperforms the corpus-in-context baseline only very modestly on a 128K token corpus, with the performance in terms of F1 being even closer than that of Subspan EM. However, we find that we are able to achieve much better performance using RAG than with the corpus-in-context methods when applying appropriate optimizations.
    \item In particular, we find that the explicit structured reasoning elicited by the use of the Justified QA strategy significantly improves performance in the RAG setting (+0.14 F1), while providing only a more moderate benefit in the corpus-in-context setting (+0.08 F1).
    \item Notably, the zero-shot Justified QA prompt achieves stronger performance than the few-shot baseline prompts, suggesting that the dataset-specific few-shot exemplars used in the baseline prompts provide little if any value compared to a clearly-worded instruction, when run on the instruction-tuned Gemini 1.5 LLMs.
    \item Applying an additional verification step for each candidate answer individually further improves performance slightly (+0.02 F1) in both the RAG and corpus-in-context settings.
    \item Despite the sometimes idiosyncratic wordings of QUEST's synthetically-generated questions, providing QUEST-specific instructions in the prompt improves performance only very modestly (+0.01 F1 on CiC, neutral on RAG).
    \item Performing answer verification directly on each of the entities whose documents were retrieved by the RAG step underperforms the use of Justified QA (-0.06 F1 vs. Justified QA + Verification, and -0.04 vs. Justified QA alone), despite involving considerably more LLM calls. This highlights the value of RAG + Justified QA as the primary drivers of performance improvements.
    \item At the same time, we can observe a noticeable precision/recall trade-off when performing RAG + Justified QA vs. RAG + answer verification alone, where the extra filtering performed by the Justified QA prompt increases precision in exchange for a drop in recall.
\end{itemize}

\subsection{Evaluation on Gemini 1.5 Flash}
\label{sect:quest_loft_128k_revised_evaluation_flash}

For results, see the leaderboard in Table~\ref{tab:leaderboard_quest_loft_128k_revised_test_flash}.

\begin{table}[htbp]
\footnotesize
\centering
\begin{tabular}{@{}l|ccccc@{}}
\toprule
Method & F1 & Precision & Recall & Accuracy & Subspan EM \\
\midrule
CiC Baseline & 0.57 & 0.66 & 0.64 & 0.26 & 0.43 \\
CiC Justified QA & 0.54 & 0.92 & 0.49 & 0.29 & 0.31 \\
CiC Justified QA + CoT & 0.60 & 0.92 & 0.54 & 0.30 & 0.32 \\
CiC Justified QA + Verification & 0.53 & 0.89 & 0.50 & 0.28 & 0.32 \\
CiC Justified QA + CoT + Verification & 0.57 & \textbf{0.96} & 0.51 & 0.28 & 0.30 \\
CiC Justified QA + Verification + QUEST & 0.56 & 0.90 & 0.53 & 0.32 & 0.36 \\
CiC Justified QA + CoT + Verification + QUEST & 0.56 & 0.94 & 0.51 & 0.29 & 0.32 \\
\midrule
RAG Baseline & 0.69 & 0.84 & 0.70 & 0.43 & 0.52 \\
RAG Justified QA & 0.71 & 0.93 & 0.68 & 0.43 & 0.48 \\
RAG Justified QA + CoT & 0.76 & 0.92 & 0.73 & \textbf{0.47} & 0.54 \\
RAG Justified QA + Verification & 0.73 & 0.92 & 0.70 & 0.46 & 0.51 \\
RAG Justified QA + CoT + Verification & 0.72 & 0.94 & 0.68 & 0.44 & 0.48 \\
RAG Justified QA + Verification + QUEST & 0.73 & 0.92 & 0.72 & 0.44 & 0.53 \\
RAG Justified QA + CoT + Verification + QUEST & 0.78 & \textbf{0.96} & 0.73 & 0.45 & 0.49 \\
\midrule
RAG + Verification & 0.55 & 0.47 & 0.92 & 0.18 & 0.83 \\
RAG + Verification (w/ CoT) & 0.74 & 0.84 & 0.79 & 0.46 & 0.63 \\
RAG + Verification + QUEST & 0.64 & 0.55 & \textbf{0.98} & 0.29 & \textbf{0.94} \\
RAG + Verification (w/ CoT) + QUEST & \textbf{0.79} & 0.81 & 0.86 & 0.46 & 0.69 \\
\bottomrule
\end{tabular}
\caption{Evaluation on QUEST-LOFT-128K-Revised with Gemini 1.5 Flash.}
\label{tab:leaderboard_quest_loft_128k_revised_test_flash}
\end{table}

From these results, we can make the following observations:
\begin{itemize}
    \item Baseline performance for Gemini 1.5 Flash is comparable to that of Gemini 1.5 Pro for RAG, but significantly worse for corpus-in-context.
    \item Inclusion of a natural language chain-of-thought step prior to structured JSON output significantly improves performance on Gemini 1.5 Flash, although it had a negligible or sometimes negative impact on Gemini 1.5 Pro.
    \item The benefits of the Justified QA approach carry over to Gemini 1.5 Flash in the RAG setting when using chain-of-thought, but only to a very limited degree in the corpus-in-context setting or when omitting chain-of-thought.
    \item More explicit instructions (e.g., the QUEST-specific instructions) had a higher impact on Gemini 1.5 Flash than on Gemini 1.5 Pro.
    \item The performance gap between Gemini 1.5 Pro and Gemini 1.5 Flash is smallest in the ``RAG + Verification (w/ CoT) + QUEST'' setting, suggesting that the importance of combining retrieval, chain-of-thought, detailed instructions, and decomposition of the task into a series of fine-grained judgments is even greater when dealing with smaller or less powerful models.
\end{itemize}

See Appendices~\ref{asect:retrieval_eval} and \ref{asect:answer_verification} for additional evaluations of the retrieval and answer verification steps in isolation.

\section{QUEST-LOFT-128K-Simple28}
\label{sect:quest_loft_128k_simple28}

\subsection{Atomic Sub-question Curation}

To better explore the degree to which the failures on QUEST-LOFT-128K-Revised are due to the ability to understand and reason about complex questions involving complex set operations vs.\ the ability to perform accurate judgments and exhaustive recall regarding the underlying atomic facts, we manually constructed a small dataset consisting of 28 atomic questions (i.e., ones that do not require set operations) which correspond to the sub-questions that an LLM would implicitly need to answer in order to reliably answer the 10 questions from the QUEST-LOFT-128K dev set. For each of the questions, we manually curate golden answers based on a comprehensive manual evaluation using the QUEST-LOFT-128K corpus, following a similar procedure to the one described in Section~\ref{sect:quest_loft_128k_revised}. We name this dataset \textit{QUEST-LOFT-128K-Simple28 dataset} and present evaluation results below.

See Appendix~\ref{asect:quest_loft_128k_simple28} for the full contents of the QUEST-LOFT-128K-Simple28 dataset.

\subsection{Evaluation}

\begin{table}[htbp]
\footnotesize
\centering
\begin{tabular}{@{}l|ccccc@{}}
\toprule
Method & F1 & Precision & Recall & Accuracy & Subspan EM \\
\midrule
CiC Baseline & 0.54 & 0.76 & 0.50 & 0.04 & 0.13 \\
CiC Justified QA & 0.62 & 0.90 & 0.53 & 0.09 & 0.09 \\
CiC Justified QA + CoT & 0.61 & 0.83 & 0.52 & 0.09 & 0.09 \\
CiC Justified QA + Verification & 0.62 & 0.91 & 0.51 & 0.09 & 0.09 \\
CiC Justified QA + CoT + Verification & 0.60 & 0.89 & 0.51 & 0.13 & 0.13 \\
CiC Justified QA + Verification + QUEST & 0.61 & 0.93 & 0.51 & 0.13 & 0.13 \\
CiC Justified QA + CoT + Verification + QUEST & 0.61 & \textbf{0.96} & 0.51 & 0.09 & 0.09 \\
\midrule
RAG Baseline & 0.66 & 0.86 & 0.60 & 0.13 & 0.13 \\
RAG Justified QA & 0.75 & 0.94 & 0.66 & 0.22 & 0.22 \\
RAG Justified QA + CoT & 0.69 & 0.90 & 0.62 & 0.17 & 0.17 \\
RAG Justified QA + Verification & 0.74 & 0.91 & 0.65 & 0.22 & 0.22 \\
RAG Justified QA + CoT + Verification & 0.72 & 0.92 & 0.64 & 0.17 & 0.17 \\
RAG Justified QA + Verification + QUEST & 0.73 & 0.90 & 0.66 & 0.22 & 0.22 \\
RAG Justified QA + CoT + Verification + QUEST & 0.76 & 0.90 & 0.69 & 0.17 & 0.22 \\
\midrule
RAG + Verification & 0.69 & 0.68 & \textbf{0.81} & 0.09 & 0.39 \\
RAG + Verification (w/ CoT) & \textbf{0.77} & 0.80 & 0.79 & \textbf{0.26} & 0.39 \\
RAG + Verification + QUEST & 0.67 & 0.68 & \textbf{0.81} & 0.13 & \textbf{0.43} \\
RAG + Verification (w/ CoT) + QUEST & 0.75 & 0.76 & 0.79 & 0.17 & 0.35 \\
\bottomrule
\end{tabular}
\caption{Evaluation on QUEST-LOFT-128K-Simple28 with Gemini 1.5 Pro.}
\label{tab:leaderboard_quest_loft_128k_simple28}
\end{table}

For results, see the leaderboard in Table \ref{tab:leaderboard_quest_loft_128k_simple28}. From these results, we can make the following observations:
\begin{itemize}
    \item While baseline precision values are comparable or higher overall (as expected) for these simpler questions compared to the more complex questions in QUEST-LOFT-128K-Revised, recall values are noticeably lower in both of the baseline corpus-in-context and RAG settings, suggesting that the models struggle to identify and process large numbers of relevant data points from the provided documents, even in the case where the questions are very simple.
    \item The benefits of the Justified QA approach carry over to the Simple28 questions in both the corpus-in-context and RAG settings. QUEST-specific instructions do not provide benefit on these simpler questions, however, which is expected, given that the most significant QUEST-specific idiosyncrasies centered primarily around the interpretation of the words ``and'' and ``or'', which do not appear in the simpler queries. The benefits of answer-by-answer verification are also smaller on the simpler questions, significantly underperforming the Justified QA approach when applied standalone, and having a neutral to negative effect when applied on top of Justified~QA.
    \item The benefit of RAG over corpus-in-context prompting is even more pronounced on these simple questions than on QUEST-LOFT-128K-Revised, both in the baseline setting (+0.12 F1) and when using Justified QA (+0.13 F1).
\end{itemize}

\section{Related Work}

Aside from the LOFT study, most papers that have explored the QUEST dataset to date have focused on the setting where the LLM is expected to answer based on its parametric knowledge, without making use of the corpus. For example, \cite{dhuliawala2023cove} present a technique called Chain-of-Verification (CoVe), in which they prompt the model to generate and answer verification questions to fact-check the individual answers from its initial response. They evaluate this technique on an easier filtered version of QUEST (containing only questions that require no logical operations) that they call ``Wiki-Category list''. In this setting, using  Llama 65B, they report improvement of precision from 0.12 to 0.22 using CoVe compared to ordinary few-shot prompting. CoVe is conceptually similar to the answer verification step that we evaluate, although their focus is on basic few-shot prompting. \cite{thirukovalluru2024atomicselfconsistency} present a technique called Atomic Self-Consistency (ASC), in which they generate multiple samples and then merge the answers lists to yield a single list of weighted candidate answers, which they then filter based on a minimum vote threshold. With this technique, using ChatGPT, they report precision of 0.123, recall of 0.104, and F1 of 0.098. \cite{lee2024craftingincontextexamplesaccording} explore the effect of the choice of few-shot exemplars and the ordering of the answers within the exemplars. They show modest improvement in performance by selecting exemplars for which the LLM knows some but not all of the correct answers, and by ordering the answers in order of decreasing perplexity. Our approach differs from these, in that we focus on how to improve the LLM's ability to retrieve and use the documents in the corpus, and on prompting approaches that yield structured outputs containing evidence and reasoning.

Approaches that evaluate retrieval strategies on QUEST include SetBERT~\citep{mai2024setbert}, which fine-tunes a BERT embedding model on examples involving set operations. They report improved retrieval recall@100 on 336M parameter models (0.314 vs. 0.300), and demonstrate comparable retrieval recall@20 on a 110M parameter fine-tuned model as on a vanilla 336M BERT model (0.145 vs. 0.146). Our approach differs from theirs, in that we focus on strategies for more effectively executing the QA task, while keeping the retrieval model fixed. Note that these numbers are not directly comparable with the recall values we report for QUEST-LOFT-128K, due to significant difference in the size of the corpora used in QUEST (325,505 entities) vs. QUEST-LOFT-128K (328 entities) and the corresponding difference in retrieval difficulty.

At the time of writing, there have not yet been any other results published on the QUEST-LOFT datasets aside from the original LOFT paper~\citep{lee2024loft}.

Our use of structured outputs containing detailed evidence and reasoning is related to the technique of chain-of-thought~\citep{wei2022cot}, in which the accuracy of an LLM's final answer is typically found to be higher if the model is made to output its step-by-step reasoning prior to outputting the final answer. Indeed, in designing the Justified QA output data structure, we took care to have the evidence and reasoning-related fields be output prior to the final judgments and final answer, so as to ensure that this chain-of-thought-like effect could be achieved. Regarding the effectiveness of JSON as an output format, past research has shown mixed results, where \cite{zhou2024self} achieve strong reasoning performance using a JSON-like output format, while \cite{tam2024let} observe a significant decline in reasoning quality when a model's output is constrained to JSON. In our experiments, we find that the ability to reason directly in JSON format depends on the model. The smaller Gemini 1.5 Flash indeed showed a noticeable improvement in performance when including a natural language chain-of-thought step prior to the JSON output, while the more powerful Gemini 1.5 Pro was able to achieve equal or better performance when outputting the JSON-based Justified QA structure directly.

Our Justified QA strategy is also closely related to the task of Attributed QA~\citep{bohnet2022attributed}, in which an LLM is expected to output a set of evidence snippets alongside the final answer. We use the term Justified QA here to indicate that the output data structure is expanded to include a more comprehensive justification of the final answers, beyond just the evidence snippets, including also information such as the model's reasoning and additional candidate answers that were considered.

Our use of an optional answer verification step is closely related to a line of research that investigates the degree to which LLMs are able to verify or correct their own outputs. Aside from the Chain-of-Verification (CoVe)~\citep{dhuliawala2023cove} technique mentioned above, there has been a large body of research into self-verification and self-correction techniques~\citep{kim2024language, shinn2024reflexion, miao2023selfcheck, gou2023critic, cohen2023lm, du2023improving}. At the same time, \cite{huang2023large} find that self-correction largely fails to improve performance on reasoning tasks, in comparison with a well-designed prompt for the initial generation step. Similarly, \cite{kamoi2024can} suggest based on a survey of literature to date that such intrinsic self-correction (i.e., self-correction without the benefit of external tools or additional knowledge) has so far been shown to be effective only in the specific case where the LLM's initial output is inherently decomposable. This special case matches, for example, the setting of QUEST, where the answer consists of a list of entity names. The results of our experiments are consistent with these findings, in that we find that the majority of the value provided by answer verification can be achieved more directly and cheaply by improving the original QA prompt to incorporate evidence and reasoning in its output, but that a small amount of additional improvement is still achievable through individual verification of the decomposed candidate answers.

The relative performance of corpus-in-context vs.\ RAG methods has been studied in several lines of work with varying results. The LOFT study described earlier~\citep{lee2024loft} found that corpus-in-context prompting yields similar accuracy to RAG with k=40 (i.e., when retrieving the top 40 passages), when evaluated on QA tasks with a corpus size of 128K tokens, with corpus-in-context outperforming when the corpus is smaller and RAG outperforming on larger corpuses. \cite{li2024retrieval} similarly found corpus-in-context prompting to yield similar performance to RAG with k=50, while significantly outperforming RAG with k=5. Our findings are compatible with both of these studies, while showing that RAG with k=40 can be optimized to significantly outperform corpus-in-context prompting when combined with improved prompt formats.

\section{Limitations}
\label{sect:limitations}

One limitation of our investigation is the small number of examples in the evaluated datasets (100 for QUEST-LOFT-128K-Revised and 28 for QUEST-LOFT-128K-Simple28), which limits the statistical significance of the conclusions that we can draw. Additionally, the small size of the dev set provided with QUEST-LOFT-128K (only 10 examples) made it infeasible to rely purely on the dev set for quality iteration signals, which increases the risk of over-fitting to the test set. As a partial mitigation for these risks, we avoided the use of automated prompt optimization techniques and focused on prompts that we considered the simplest and most intuitive ways of illustrating a given technique. In cases where we had evaluated multiple prompt variants that differed only in formatting or style (e.g., as part of the process of simplifying or improving prompt consistency), we verified informally that the qualitative conclusions that we report were observable on average across the prompt variants. In order to verify these conclusions more robustly, it will be important in the future to extend this line of experiments to larger eval sets drawn from a more diverse range of QA tasks.

\section{Conclusion}
\label{sect:conclusion}

In this paper, we analyzed the factors contributing to current poor performance on QUEST-LOFT, including a comprehensive human evaluation leading to a revised set of golden answers. We evaluated a range of techniques, including chain-of-thought reasoning and multi-step prompting techniques. Through a combination of these techniques, we improved the SOTA on QUEST-LOFT-128K-Revised by a significant margin (F1 +0.16, accuracy +0.16, subspan EM +0.11). The results of our experiments suggest a number of promising directions for future work. In particular, we see a strong need for continued investment in retrieval-based (i.e., RAG) strategies, despite the increasing availability of long-context LLMs. We observe that the choice of prompting strategy used for the QA step can make a significant difference in performance, with long-form outputs that expose the LLM's detailed reasoning steps contributing to improvements in both precision and recall, in addition to making the results more interpretable. When using the smaller Gemini 1.5 Flash model, we further observe a significant gap between the performance that is achievable when an LLM is made to perform a series of targeted and independent judgments about specific candidate answers based on just the documents relevant to that answer, compared to when the LLM is asked
to directly perform the full sequence of reasoning for determining the entire answer set in a single prompt. One challenge for future research will be to find ways to maintain this top performance while managing costs as the number of questions and documents increase. We are also interested in expanding our evaluation to a wider range of corpus-based QA benchmarks, including ones that require aggregation of information about the same entity from multiple different documents, more diverse patterns of multi-hop reasoning, or generation of long-form answers with fact attribution. At the same time, we see a need for further investment in high-quality manually-curated evaluation datasets that reflect more fully the challenges of real-world tasks, while aiming for the high quality bar of manual curation that can be achieved at the relatively small scale of a few hundred examples and a correspondingly small corpus.

\section{Acknowledgements}

We thank Ilya Tolstikhin, Afroz Siddiqui, Sören Meyer-Eppler, Hylke Buisman, Brian Rosen, and John Palowitch for contributions to the codebase used in this research. We thank Jinhyuk Lee and Iftekhar Naim for guidance in the use of Gecko embedding-based retrieval and in reproducing the baseline experiments from the LOFT paper. We thank Peter Shaw for guidance on the QUEST dataset. We thank Daniel Keysers, Chuck Sugnet, Andrii Maksai, and Jeremiah Harmsen for helpful feedback on the paper.

\bibliography{main}

\begin{thebibliography}{27}
\providecommand{\natexlab}[1]{#1}
\providecommand{\url}[1]{\texttt{#1}}
\expandafter\ifx\csname urlstyle\endcsname\relax
  \providecommand{\doi}[1]{doi: #1}\else
  \providecommand{\doi}{doi: \begingroup \urlstyle{rm}\Url}\fi

\bibitem[Bohnet et~al.(2022)Bohnet, Tran, Verga, Aharoni, Andor, Soares,
  Ciaramita, Eisenstein, Ganchev, Herzig, et~al.]{bohnet2022attributed}
B.~Bohnet, V.~Q. Tran, P.~Verga, R.~Aharoni, D.~Andor, L.~B. Soares,
  M.~Ciaramita, J.~Eisenstein, K.~Ganchev, J.~Herzig, et~al.
\newblock Attributed question answering: Evaluation and modeling for attributed
  large language models.
\newblock \emph{arXiv preprint arXiv:2212.08037}, 2022.

\bibitem[Bousquet et~al.(2024)Bousquet, Scales, Sch{\"a}rli, and
  Tolstikhin]{onetwo2024github}
O.~Bousquet, N.~Scales, N.~Sch{\"a}rli, and I.~Tolstikhin.
\newblock {O}ne{T}wo: {I}nteracting with {L}arge {M}odels, 2024.
\newblock URL \url{https://github.com/google-deepmind/onetwo}.

\bibitem[Cohen et~al.(2023)Cohen, Hamri, Geva, and Globerson]{cohen2023lm}
R.~Cohen, M.~Hamri, M.~Geva, and A.~Globerson.
\newblock {LM} vs {LM}: Detecting factual errors via cross examination.
\newblock \emph{arXiv preprint arXiv:2305.13281}, 2023.

\bibitem[Dhuliawala et~al.(2023)Dhuliawala, Komeili, Xu, Raileanu, Li,
  Celikyilmaz, and Weston]{dhuliawala2023cove}
S.~Dhuliawala, M.~Komeili, J.~Xu, R.~Raileanu, X.~Li, A.~Celikyilmaz, and
  J.~Weston.
\newblock Chain-of-verification reduces hallucination in large language models,
  2023.
\newblock URL \url{https://arxiv.org/abs/2309.11495}.

\bibitem[Du et~al.(2023)Du, Li, Torralba, Tenenbaum, and
  Mordatch]{du2023improving}
Y.~Du, S.~Li, A.~Torralba, J.~B. Tenenbaum, and I.~Mordatch.
\newblock Improving factuality and reasoning in language models through
  multiagent debate.
\newblock \emph{arXiv preprint arXiv:2305.14325}, 2023.

\bibitem[Gao et~al.(2023)Gao, Xiong, Gao, Jia, Pan, Bi, Dai, Sun, Wang, and
  Wang]{gao2023rag}
Y.~Gao, Y.~Xiong, X.~Gao, K.~Jia, J.~Pan, Y.~Bi, Y.~Dai, J.~Sun, M.~Wang, and
  H.~Wang.
\newblock Retrieval-augmented generation for large language models: A survey.
\newblock \emph{arXiv preprint arXiv:2312.10997}, 2023.

\bibitem[Gou et~al.(2023)Gou, Shao, Gong, Shen, Yang, Duan, and
  Chen]{gou2023critic}
Z.~Gou, Z.~Shao, Y.~Gong, Y.~Shen, Y.~Yang, N.~Duan, and W.~Chen.
\newblock Critic: Large language models can self-correct with tool-interactive
  critiquing.
\newblock \emph{arXiv preprint arXiv:2305.11738}, 2023.

\bibitem[Huang et~al.(2023{\natexlab{a}})Huang, Chen, Mishra, Zheng, Yu, Song,
  and Zhou]{huang2023large}
J.~Huang, X.~Chen, S.~Mishra, H.~S. Zheng, A.~W. Yu, X.~Song, and D.~Zhou.
\newblock Large language models cannot self-correct reasoning yet.
\newblock \emph{arXiv preprint arXiv:2310.01798}, 2023{\natexlab{a}}.

\bibitem[Huang et~al.(2023{\natexlab{b}})Huang, Yu, Ma, Zhong, Feng, Wang,
  Chen, Peng, Feng, Qin, et~al.]{huang2023survey}
L.~Huang, W.~Yu, W.~Ma, W.~Zhong, Z.~Feng, H.~Wang, Q.~Chen, W.~Peng, X.~Feng,
  B.~Qin, et~al.
\newblock A survey on hallucination in large language models: Principles,
  taxonomy, challenges, and open questions.
\newblock \emph{arXiv preprint arXiv:2311.05232}, 2023{\natexlab{b}}.

\bibitem[Kamoi et~al.(2024)Kamoi, Zhang, Zhang, Han, and Zhang]{kamoi2024can}
R.~Kamoi, Y.~Zhang, N.~Zhang, J.~Han, and R.~Zhang.
\newblock When can {LLMs} actually correct their own mistakes? a critical
  survey of self-correction of {LLMs}.
\newblock \emph{Transactions of the Association for Computational Linguistics},
  12:\penalty0 1417--1440, 2024.

\bibitem[Kim et~al.(2024)Kim, Baldi, and McAleer]{kim2024language}
G.~Kim, P.~Baldi, and S.~McAleer.
\newblock Language models can solve computer tasks.
\newblock \emph{Advances in Neural Information Processing Systems}, 36, 2024.

\bibitem[Lee et~al.(2024{\natexlab{a}})Lee, Chen, Dai, Dua, Sachan, Boratko,
  Luan, Arnold, Perot, Dalmia, Hu, Lin, Pasupat, Amini, Cole, Riedel, Naim,
  Chang, and Guu]{lee2024loft}
J.~Lee, A.~Chen, Z.~Dai, D.~Dua, D.~S. Sachan, M.~Boratko, Y.~Luan, S.~M.~R.
  Arnold, V.~Perot, S.~Dalmia, H.~Hu, X.~Lin, P.~Pasupat, A.~Amini, J.~R. Cole,
  S.~Riedel, I.~Naim, M.-W. Chang, and K.~Guu.
\newblock Can long-context language models subsume retrieval, {RAG}, {SQL}, and
  more?, 2024{\natexlab{a}}.
\newblock URL \url{https://arxiv.org/abs/2406.13121}.

\bibitem[Lee et~al.(2024{\natexlab{b}})Lee, Dai, Ren, Chen, Cer, Cole, Hui,
  Boratko, Kapadia, Ding, Luan, Duddu, Abrego, Shi, Gupta, Kusupati, Jain,
  Jonnalagadda, Chang, and Naim]{lee2024gecko}
J.~Lee, Z.~Dai, X.~Ren, B.~Chen, D.~Cer, J.~R. Cole, K.~Hui, M.~Boratko,
  R.~Kapadia, W.~Ding, Y.~Luan, S.~M.~K. Duddu, G.~H. Abrego, W.~Shi, N.~Gupta,
  A.~Kusupati, P.~Jain, S.~R. Jonnalagadda, M.-W. Chang, and I.~Naim.
\newblock Gecko: Versatile text embeddings distilled from large language
  models, 2024{\natexlab{b}}.
\newblock URL \url{https://arxiv.org/abs/2403.20327}.

\bibitem[Lee et~al.(2024{\natexlab{c}})Lee, Atreya, Ye, and
  Choi]{lee2024craftingincontextexamplesaccording}
Y.~Lee, P.~Atreya, X.~Ye, and E.~Choi.
\newblock Crafting in-context examples according to {LMs}' parametric
  knowledge, 2024{\natexlab{c}}.
\newblock URL \url{https://arxiv.org/abs/2311.09579}.

\bibitem[Lewis et~al.(2020)Lewis, Perez, Piktus, Petroni, Karpukhin, Goyal,
  K{\"u}ttler, Lewis, Yih, Rockt{\"a}schel, et~al.]{lewis2020rag}
P.~Lewis, E.~Perez, A.~Piktus, F.~Petroni, V.~Karpukhin, N.~Goyal,
  H.~K{\"u}ttler, M.~Lewis, W.-t. Yih, T.~Rockt{\"a}schel, et~al.
\newblock Retrieval-augmented generation for knowledge-intensive nlp tasks.
\newblock \emph{Advances in Neural Information Processing Systems},
  33:\penalty0 9459--9474, 2020.

\bibitem[Li et~al.(2024)Li, Li, Zhang, Mei, and Bendersky]{li2024retrieval}
Z.~Li, C.~Li, M.~Zhang, Q.~Mei, and M.~Bendersky.
\newblock Retrieval augmented generation or long-context {LLMs}? a
  comprehensive study and hybrid approach.
\newblock In \emph{Proceedings of the 2024 Conference on Empirical Methods in
  Natural Language Processing: Industry Track}, pages 881--893, 2024.

\bibitem[Mai et~al.(2024)Mai, Gauch, and Adams]{mai2024setbert}
Q.~Mai, S.~Gauch, and D.~Adams.
\newblock Setbert: Enhancing retrieval performance for boolean logic and set
  operation queries, 2024.
\newblock URL \url{https://arxiv.org/abs/2406.17282}.

\bibitem[Malaviya et~al.(2023)Malaviya, Shaw, Chang, Lee, and
  Toutanova]{malaviya2023quest}
C.~Malaviya, P.~Shaw, M.-W. Chang, K.~Lee, and K.~Toutanova.
\newblock Quest: A retrieval dataset of entity-seeking queries with implicit
  set operations, 2023.
\newblock URL \url{https://arxiv.org/abs/2305.11694}.

\bibitem[Miao et~al.(2023)Miao, Teh, and Rainforth]{miao2023selfcheck}
N.~Miao, Y.~W. Teh, and T.~Rainforth.
\newblock {SelfCheck}: Using {LLMs} to zero-shot check their own step-by-step
  reasoning.
\newblock \emph{arXiv preprint arXiv:2308.00436}, 2023.

\bibitem[Min et~al.(2021)Min, Lee, Chang, Toutanova, and
  Hajishirzi]{min2021joint}
S.~Min, K.~Lee, M.-W. Chang, K.~Toutanova, and H.~Hajishirzi.
\newblock Joint passage ranking for diverse multi-answer retrieval.
\newblock \emph{arXiv preprint arXiv:2104.08445}, 2021.

\bibitem[Min et~al.(2023)Min, Krishna, Lyu, Lewis, Yih, Koh, Iyyer,
  Zettlemoyer, and Hajishirzi]{min2023factscore}
S.~Min, K.~Krishna, X.~Lyu, M.~Lewis, W.-t. Yih, P.~W. Koh, M.~Iyyer,
  L.~Zettlemoyer, and H.~Hajishirzi.
\newblock Factscore: Fine-grained atomic evaluation of factual precision in
  long form text generation.
\newblock \emph{arXiv preprint arXiv:2305.14251}, 2023.

\bibitem[Shinn et~al.(2024)Shinn, Cassano, Gopinath, Narasimhan, and
  Yao]{shinn2024reflexion}
N.~Shinn, F.~Cassano, A.~Gopinath, K.~Narasimhan, and S.~Yao.
\newblock Reflexion: Language agents with verbal reinforcement learning.
\newblock \emph{Advances in Neural Information Processing Systems}, 36, 2024.

\bibitem[Tam et~al.(2024)Tam, Wu, Tsai, Lin, Lee, and Chen]{tam2024let}
Z.~R. Tam, C.-K. Wu, Y.-L. Tsai, C.-Y. Lin, H.-y. Lee, and Y.-N. Chen.
\newblock Let me speak freely? {A} study on the impact of format restrictions
  on performance of large language models.
\newblock \emph{arXiv preprint arXiv:2408.02442}, 2024.

\bibitem[Team et~al.(2024)Team, Georgiev, Lei, Burnell, Bai, Gulati, Tanzer,
  Vincent, Pan, Wang, et~al.]{team2024gemini}
G.~Team, P.~Georgiev, V.~I. Lei, R.~Burnell, L.~Bai, A.~Gulati, G.~Tanzer,
  D.~Vincent, Z.~Pan, S.~Wang, et~al.
\newblock Gemini 1.5: Unlocking multimodal understanding across millions of
  tokens of context.
\newblock \emph{arXiv preprint arXiv:2403.05530}, 2024.

\bibitem[Thirukovalluru et~al.(2024)Thirukovalluru, Huang, and
  Dhingra]{thirukovalluru2024atomicselfconsistency}
R.~Thirukovalluru, Y.~Huang, and B.~Dhingra.
\newblock Atomic self-consistency for better long form generations, 2024.
\newblock URL \url{https://arxiv.org/abs/2405.13131}.

\bibitem[Wei et~al.(2022)Wei, Wang, Schuurmans, Bosma, Xia, Chi, Le, Zhou,
  et~al.]{wei2022cot}
J.~Wei, X.~Wang, D.~Schuurmans, M.~Bosma, F.~Xia, E.~Chi, Q.~V. Le, D.~Zhou,
  et~al.
\newblock Chain-of-thought prompting elicits reasoning in large language
  models.
\newblock \emph{Advances in neural information processing systems},
  35:\penalty0 24824--24837, 2022.

\bibitem[Zhou et~al.(2024)Zhou, Pujara, Ren, Chen, Cheng, Le, Chi, Zhou,
  Mishra, and Zheng]{zhou2024self}
P.~Zhou, J.~Pujara, X.~Ren, X.~Chen, H.-T. Cheng, Q.~V. Le, E.~H. Chi, D.~Zhou,
  S.~Mishra, and H.~S. Zheng.
\newblock Self-discover: Large language models self-compose reasoning
  structures.
\newblock \emph{arXiv preprint arXiv:2402.03620}, 2024.

\end{thebibliography}

\clearpage
\appendix
\section*{Appendix}

\section{Retrieval eval}
\label{asect:retrieval_eval}

Although in our evaluations up till now we focused on accuracy or F1 of the ``answers'' to the QA task, we can alternatively view the QA strategies as another form of document retrieval. Specifically, the Justified QA strategy can be thought of as filtering an original set of documents down to just the documents that it cited as evidence for its candidate answers. Similarly, the CiC and RAG baseline prompts can be thought of as filtering the documents down to those that correspond to the retrieved answers. Based on this interpretation, we conduct an investigation of the effect of each of the strategies on retrieval recall. The results can be seen in Tables~\ref{tab:leaderboard_retrieval_quest_loft_128k_revised_test_pro}, \ref{tab:leaderboard_retrieval_quest_loft_128k_revised_test_flash} and \ref{tab:leaderboard_retrieval_quest_loft_128k_simple28}.

The metrics that we track are defined as follows:

\begin{itemize}
    \item \textbf{Recall@K:} The fraction of golden answers (i.e., answers rated as MATCH) whose corresponding document appears among the top K results returned by the given retrieval method.
    \item \textbf{MRecall@K:} Metric introduced in \citep{min2021joint} and used for multi-document retrieval evaluations in the LOFT study~\citep{lee2024loft}. Gives a score of 1.0 if all of the golden answers (i.e., answers rated as MATCH), or at least K of them, have their corresponding document appearing among the top K results returned by the given retrieval method.
\end{itemize}

\begin{table}[ht]
\centering
\begin{tabular}{@{}l|cccc@{}}
\toprule
Method & MRecall@3 & Recall@20 & Recall@40 & Recall@100 \\
\midrule
Naive retrieval (first K docs in corpus) & 0.12 & 0.17 & 0.24 & 0.39 \\
Embedding-based retrieval & 0.67 & \textbf{0.95} & \textbf{0.99} & \textbf{1.00} \\
\midrule
CiC Baseline & 0.52 & 0.70 & 0.70 & 0.70 \\
CiC Justified & 0.60 & 0.76 & 0.76 & 0.76 \\
CiC Justified - CoT & 0.55 & 0.76 & 0.76 & 0.76 \\
CiC Justified - QUEST & 0.61 & 0.79 & 0.79 & 0.79 \\
CiC Justified - CoT - QUEST & 0.58 & 0.77 & 0.77 & 0.77 \\
\midrule
RAG Baseline (RaR) & 0.58 & 0.72 & 0.72 & 0.72 \\
RAG Justified & 0.76 & 0.86 & 0.86 & 0.86 \\
RAG Justified - CoT & 0.69 & 0.83 & 0.83 & 0.83 \\
RAG Justified - QUEST & \textbf{0.79} & 0.89 & 0.89 & 0.89 \\
RAG Justified - CoT - QUEST & 0.71 & 0.82 & 0.82 & 0.82 \\
\bottomrule
\end{tabular}
\caption{Retrieval evaluation on QUEST-LOFT-128K-Revised with Gemini 1.5 Pro.}
\label{tab:leaderboard_retrieval_quest_loft_128k_revised_test_pro}
\end{table}

\begin{table}[ht]
\centering
\begin{tabular}{@{}l|cccc@{}}
\toprule
Method & MRecall@3 & Recall@20 & Recall@40 & Recall@100 \\
\midrule
Naive retrieval (first K docs in corpus) & 0.12 & 0.17 & 0.24 & 0.39 \\
Embedding-based retrieval & \textbf{0.67} & \textbf{0.95} & \textbf{0.99} & \textbf{1.00} \\
\midrule
CiC Baseline & 0.43 & 0.64 & 0.64 & 0.64 \\
CiC Justified & 0.34 & 0.53 & 0.54 & 0.55 \\
CiC Justified - CoT & 0.33 & 0.57 & 0.57 & 0.57 \\
CiC Justified - QUEST & 0.37 & 0.55 & 0.55 & 0.57 \\
CiC Justified - CoT - QUEST & 0.37 & 0.56 & 0.56 & 0.56 \\
\midrule
RAG Baseline (RaR) & 0.53 & 0.70 & 0.70 & 0.70 \\
RAG Justified & 0.59 & 0.76 & 0.76 & 0.76 \\
RAG Justified - CoT & 0.65 & 0.78 & 0.78 & 0.78 \\
RAG Justified - QUEST & 0.60 & 0.77 & 0.77 & 0.77 \\
RAG Justified - CoT - QUEST & 0.62 & 0.79 & 0.79 & 0.79 \\
\bottomrule
\end{tabular}
\caption{Retrieval evaluation on QUEST-LOFT-128K-Revised with Gemini 1.5 Flash.}
\label{tab:leaderboard_retrieval_quest_loft_128k_revised_test_flash}
\end{table}

\begin{table}[ht]
\centering
\begin{tabular}{@{}l|cccc@{}}
\toprule
Method & MRecall@3 & Recall@20 & Recall@40 & Recall@100 \\
\midrule
Naive retrieval (first K docs in corpus) & 0.04 & 0.12 & 0.21 & 0.43 \\
Embedding-based retrieval & \textbf{0.61} & \textbf{0.74} & \textbf{0.90} & \textbf{0.98} \\
\midrule
CiC Baseline & 0.35 & 0.50 & 0.50 & 0.50 \\
CiC Justified & 0.39 & 0.53 & 0.53 & 0.53 \\
CiC Justified - CoT & 0.39 & 0.53 & 0.53 & 0.53 \\
CiC Justified - QUEST & 0.43 & 0.55 & 0.55 & 0.55 \\
CiC Justified - CoT - QUEST & 0.35 & 0.53 & 0.54 & 0.56 \\
\midrule
RAG Baseline (RaR) & 0.52 & 0.59 & 0.60 & 0.60 \\
RAG Justified & 0.57 & 0.64 & 0.67 & 0.67 \\
RAG Justified - CoT & \textbf{0.61} & 0.65 & 0.66 & 0.66 \\
RAG Justified - QUEST & 0.57 & 0.66 & 0.69 & 0.69 \\
RAG Justified - CoT - QUEST & \textbf{0.61} & 0.71 & 0.74 & 0.74 \\
\bottomrule
\end{tabular}
\caption{Retrieval evaluation on QUEST-LOFT-128K-Simple28 with Gemini 1.5 Pro.}
\label{tab:leaderboard_retrieval_quest_loft_128k_simple28}
\end{table}

From these results, we can make the following observations:
\begin{itemize}
    \item For this relatively small corpus, the Recall@40 for embedding-based retrieval is essentially perfect, which explains why the RAG-based QA strategies would be expected to perform at least as well as the corpus-in-context strategies. I.e., for this corpus, there is no drawback and essentially only benefit in including an embedding-based retrieval step prior to prompting the LLM to further filter the results.
    \item The Justified QA prompt is noticeably more effective than the baseline prompts both at preserving overall recall (as seen in Recall@40) and in promoting correct candidates to the top of the list (as seen in MRecall@3).
    \item Inclusion of the QUEST-specific instruction in the Justified QA prompt surprisingly yields a noticeable improvement in recall, even for the Simple28 dataset, despite the end-to-end QA strategies showing at most modest improvements in recall and in some cases drops in overall F1 when including the QUEST-specific instruction. This suggests that this positive effect of the QUEST-specific instruction is being offset through some combination of negative effects on precision and/or a lack of corresponding recall benefit in the answer verification step.
\end{itemize}

\section{Answer verification eval}
\label{asect:answer_verification_eval}

In order to isolate the effect of the answer verification component, we evaluate this component in isolation on a pair of binary classification datasets that we derive from QUEST-LOFT-128K-Revised and QUEST-LOFT-128K-Simple28, respectively, using the individual entities from the golden answers (as positive examples) and a selection of other entities that were incorrectly included in answers proposed by the LLM in earlier eval runs (as negative examples). The answer verification dataset for QUEST-LOFT-128K-Revised contains a total of 371 examples, while the one for QUEST-LOFT-128K-Simple28 contains 453 examples (the larger number of examples stemming from the significantly larger answer sets associated with the questions in QUEST-LOFT-128K-Simple28). The results can be seen in Tables~\ref{tab:leaderboard_answer_verification_quest_loft_128k_revised_test_pro}, \ref{tab:leaderboard_answer_verification_quest_loft_128k_revised_test_flash} and \ref{tab:leaderboard_answer_verification_quest_loft_128k_simple28}.

\begin{table}[ht]
\centering
\begin{tabular}{@{}l|cccc@{}}
\toprule
Method & Precision & Recall & Accuracy & F1 \\
\midrule
Basic & 0.84 & \textbf{0.89} & 0.83 & 0.86 \\
CoT & 0.90 & 0.81 & 0.83 & 0.85 \\
QUEST & 0.87 & \textbf{0.89} & 0.85 & \textbf{0.88} \\
CoT - QUEST & \textbf{0.91} & 0.86 & \textbf{0.86} & \textbf{0.88} \\
\bottomrule
\end{tabular}
\caption{Answer verification evaluation on QUEST-LOFT-128K-Revised with Gemini 1.5 Pro.}
\label{tab:leaderboard_answer_verification_quest_loft_128k_revised_test_pro}
\end{table}

\begin{table}[ht]
\centering
\begin{tabular}{@{}l|cccc@{}}
\toprule
Method & Precision & Recall & Accuracy & F1 \\
\midrule
Basic & 0.79 & 0.93 & 0.80 & 0.85 \\
CoT & \textbf{0.90} & 0.77 & 0.80 & 0.83 \\
QUEST & 0.80 & \textbf{0.96} & \textbf{0.83} & \textbf{0.87} \\
CoT - QUEST & 0.85 & 0.84 & 0.80 & 0.84 \\
\bottomrule
\end{tabular}
\caption{Answer verification evaluation on QUEST-LOFT-128K-Revised with Gemini 1.5 Flash.}
\label{tab:leaderboard_answer_verification_quest_loft_128k_revised_test_flash}
\end{table}

\begin{table}[ht]
\centering
\begin{tabular}{@{}l|cccc@{}}
\toprule
Method & Precision & Recall & Accuracy & F1 \\
\midrule
Basic & 0.75 & 0.89 & 0.75 & 0.81 \\
CoT & \textbf{0.79} & 0.87 & \textbf{0.78} & \textbf{0.83} \\
QUEST & 0.73 & \textbf{0.92} & 0.73 & 0.81 \\
CoT - QUEST & 0.76 & 0.90 & 0.76 & 0.82 \\
\bottomrule
\end{tabular}
\caption{Answer verification evaluation on QUEST-LOFT-128K-Simple28 with Gemini 1.5 Pro.}
\label{tab:leaderboard_answer_verification_quest_loft_128k_simple28}
\end{table}

From these results, we can make the following observations:
\begin{itemize}
    \item Including QUEST-specific instructions in the answer verification prompt does improve recall modestly, but to a much lesser degree than in the Justified QA prompt.
    \item Overall, it appears that the modest precision of the answer verification step is likely one of the main limiting factors in its effectiveness. The QUEST-specific instruction's effect of improving answer verification precision on the QUEST-LOFT-128K-Revised dataset while hurting precision on QUEST-LOFT-128K-Simple28 aligns correspondingly to the effect of the QUEST-specific instructions on the end-to-end QA performance seen in Sections~\ref{sect:quest_loft_128k_revised} and~\ref{sect:quest_loft_128k_simple28}.
\end{itemize}

\section{QA Strategies}
\label{asect:qa_strategies}

In this section, we provide details on the prompts used in each of the QA strategies.

\subsection{Corpus-in-context (CiC) QA}
\label{asect:cic_qa}

See \cite{lee2024loft} for details.

\subsection{Retrieve-and-read (RaR) QA}
\label{asect:rar_qa}

See \cite{lee2024loft} for details.

\subsection{Justified QA}
\label{asect:verifiable_qa}

\subsubsection{Justified QA (default)}
\label{asect:verifiable_qa_default}

\begin{lstlisting}[
    breaklines=true,
    basicstyle=\footnotesize\ttfamily,
    literate={```}{\textasciigrave\textasciigrave\textasciigrave}{3}
]
Your task is to answer a given question based on the provided documents.

Please do so by outputting a response in the form of a JSON object conforming to the following schema:

```json
{
  "question": "<Original Question>",
  "candidate_answers": [
    {
      "candidate_answer": "<Candidate Answer>",
      "evidence_for": [{"doc_id": "<Doc ID>", "text": "<Snippets separated by ...>"}],
      "evidence_against": [{"doc_id": "<Doc ID>", "text": "<Snippets separated by ...>"}],
      "reasoning": "<Detailed reasoning addressing each of the relevant criteria>",
      "final_judgment": "<TRUE/FALSE>"
    }
  ],
  "answer": ["<List of correct answers based *only* on TRUE candidate answers>"],
  "answer_doc_ids": ["<List of doc_ids of the documents containing the key evidence for the correct answers"],
}
```

Detailed instructions:
* Please prioritize recall in the stage of gathering candidate answers, and then filter the candidate answers for precision via the "reasoning" and "final_judgment" fields.
* For questions that are expected to return a list of values, each individual value should be treated as a separate "candidate answer".
* In the text snippets provided as evidence for/against, please include **every** sentence (or portion of sentence) from the original document that contains information relevant to the question, but skipping irrelevant sentences (inserting "..." in their place).

===== Documents =====
{{documents}}

===== Question =====
{{question}}
\end{lstlisting}

\subsubsection{Justified QA - CoT}
\label{asect:verifiable_qa_cot}

\begin{lstlisting}[
    breaklines=true,
    basicstyle=\footnotesize\ttfamily,
    literate={```}{\textasciigrave\textasciigrave\textasciigrave}{3}
]
Your task is to answer a given question based on the provided documents.

Please first take some time to take notes, organize your thoughts, and brainstorm possible answers (including borderline ones). Then output a final response in the form of a JSON list conforming to the following schema.

```json
{
  "type": "VerifiableContextualQAResponse",
  "question": "<Original Question>",
  "candidate_answers": [
    {
      "candidate_answer": "<Candidate Answer>",
      "evidence_for": [{"doc_id": "<Doc ID>", "text": "<Snippets separated by ...>"}],
      "evidence_against": [{"doc_id": "<Doc ID>", "text": "<Snippets separated by ...>"}],
      "reasoning": "<Detailed reasoning addressing each of the relevant criteria>",
      "final_judgment": "<TRUE/FALSE>"
    }
  ],
  "answer": ["<List of correct answers based *only* on TRUE candidate answers>"],
  "answer_doc_ids": ["<List of doc_ids of the documents containing the key evidence for the correct answers"],
}
```

Detailed instructions:
* Please prioritize recall in the stage of gathering candidate answers, and then filter the candidate answers for precision via the "reasoning" and "final_judgment" fields.
* For questions that are expected to return a list of values, each individual value should be treated as a separate "candidate answer".
* In the text snippets provided as evidence for/against, please include **every** sentence (or portion of sentence) from the original document that contains information relevant to the question, but skipping irrelevant sentences (inserting "..." in their place).

Please output each of the above steps in order, demarcated with a header like the following:
===== Step 1: Notes =====
===== Step 2: JSON response =====
===== END =====

===== Documents =====
{{documents}}

===== Question =====
{{question}}
\end{lstlisting}

\subsubsection{Justified QA - QUEST}
\label{asect:verifiable_qa_quest}

Same as the default prompt, but with the following additional QUEST-specific bullet point added to the ``Detailed instructions'' section:

\begin{lstlisting}[
    breaklines=true,
    basicstyle=\footnotesize\ttfamily,
    literate={```}{\textasciigrave\textasciigrave\textasciigrave}{3}
]
* Note that the question comes from a dataset that was generated semi-automatically based on names of Wikipedia categories, combined with 'and' to mean set intersection and 'or' to mean set union, and the answer is expected to be a list of entity names (i.e., document titles). Please interpret the question accordingly.
\end{lstlisting}

\subsubsection{Justified QA - CoT - QUEST}
\label{asect:verifiable_qa_cot_quest}

Same as the CoT prompt, but with the additional QUEST-specific bullet point added to the ``Detailed instructions'' section.

\subsubsection{Document formatting}
\label{asect:document_formatting}

In all of the above prompt variants, the documents are represented in the same format as in the baseline strategies, where each document is formatted using the template shown below and printed one after the other:

\begin{lstlisting}[
    breaklines=true,
    basicstyle=\footnotesize\ttfamily,
    literate={```}{\textasciigrave\textasciigrave\textasciigrave}{3}
]
ID: {{doc_id}} | TITLE: {{title}} | CONTENT: {{text}}
\end{lstlisting}

Note that while \textit{doc\_id} and \textit{title} are typically short strings (without newlines), the document's \textit{text} typically spans many lines.

\section{Answer verification}
\label{asect:answer_verification}

\subsection{Answer verification - Generic}
\label{asect:answer_verification_generic}

\begin{lstlisting}[
    breaklines=true,
    basicstyle=\footnotesize\ttfamily,
    literate={```}{\textasciigrave\textasciigrave\textasciigrave}{3}
]
Your task is to judge whether a given candidate answer is a valid answer to the given question based on the provided documents.

Please do so by outputting a response in the form of a JSON object conforming to the following schema.

```json
{
  "candidate_answer": "<Candidate Answer>",
  "evidence_for": [{"doc_id": "<Doc ID>", "text": "<Snippets separated by ...>"}],
  "evidence_against": [{"doc_id": "<Doc ID>", "text": "<Snippets separated by ...>"}],
  "reasoning": "<Detailed reasoning addressing each of the relevant criteria>",
  "final_judgment": "<TRUE/FALSE>"
}
```

Detailed instructions:
* In the text snippets provided as evidence for/against, please include **every** sentence (or portion of sentence) from the original document that contains information relevant to the question, but skipping irrelevant sentences (inserting "..." in their place).

===== Documents =====
{{documents}}

===== Question =====
{{question}}

===== Candidate Answer =====
{{candidate_answer}}
\end{lstlisting}

\newpage 

\subsection{Answer verification - CoT}
\label{asect:answer_verification_cot}

\begin{lstlisting}[
    breaklines=true,
    basicstyle=\footnotesize\ttfamily,
    literate={```}{\textasciigrave\textasciigrave\textasciigrave}{3}
]
Your task is to judge whether a given candidate answer is a valid answer to the given question based on the provided documents.

Please first take some time to take notes and organize your thoughts. Then after that output a final response in the form of a JSON object conforming to the following schema.

```json
{
  "candidate_answer": "<Candidate Answer>",
  "evidence_for": [{"doc_id": "<Doc ID>", "text": "<Snippets separated by ...>"}],
  "evidence_against": [{"doc_id": "<Doc ID>", "text": "<Snippets separated by ...>"}],
  "reasoning": "<Detailed reasoning addressing each of the relevant criteria>",
  "final_judgment": "<TRUE/FALSE>"
}
```

Detailed instructions:
* In the text snippets provided as evidence for/against, please include **every** sentence (or portion of sentence) from the original document that contains information relevant to the question, but skipping irrelevant sentences (inserting "..." in their place).

Please output each of the above steps in order, demarcated with a header like the following:
===== Step 1: Notes =====
===== Step 2: JSON response =====
===== END =====

===== Documents =====
{{documents}}

===== Question =====
{{question}}

===== Candidate Answer =====
{{candidate_answer}}
\end{lstlisting}

\subsection{Answer verification - QUEST}
\label{asect:answer_verification_quest}

Same as the generic prompt, but with the following additional bullet point added to the ``Detailed instructions'' section:

\begin{lstlisting}[
    breaklines=true,
    basicstyle=\footnotesize\ttfamily,
    literate={```}{\textasciigrave\textasciigrave\textasciigrave}{3}
]
* Note that the question comes from a dataset that was generated semi-automatically based on names of Wikipedia categories, combined with 'and' to mean set intersection and 'or' to mean set union, and the answer is expected to be a list of entity names (i.e., document titles). Please interpret the question accordingly.
\end{lstlisting}

\subsection{Answer verification - CoT - QUEST}
\label{asect:answer_verification_cot_quest}

Same as the CoT prompt, but with the additional QUEST-specific bullet point added to the ``Detailed instructions'' section.

\section{QUEST-LOFT-128K-Revised}
\label{asect:quest_loft_128k_revised}

Below are the details of the manual corrections made to the golden answer sets of QUEST-LOFT-128K in the creation of QUEST-LOFT-128K-Revised.

As described in Section~\ref{sect:quest_loft_128k_revised_curation}, we rate answers by classifying them into the following three buckets:
\begin{itemize}
\item \textbf{MATCH:} The answer clearly matches the question. It will be treated as a recall error if this answer is not included in the predicted list.
\item \textbf{DEBATABLE:} There are good arguments for and against considering the answer a match for the question. Whether this answer is included in the predicted list or not will have no effect on any of the metrics.
\item \textbf{NO\_MATCH:} The answer clearly does not match the question. It will be treated as a precision error if this answer is included in the predicted list. Note that any answer that is not rated as MATCH or DEBATABLE is in any case implicitly considered to be NO\_MATCH. We only list the NO\_MATCH answers here in the case that they were listed as a golden answer in the original QUEST dataset.
\end{itemize}

Questions with their revised answer ratings:
\begin{enumerate}
\footnotesize
\item \textbf{Novels about families set in Boston and New England.}
  \begin{itemize}
  \item \textbf{MATCH:} ``The Scarlet Letter''
  \item \textbf{DEBATABLE:} ``A Case of Need''
  \end{itemize}
\item \textbf{2009 Indian, or by Chetan Bhagat, novels}
  \begin{itemize}
  \item \textbf{MATCH:} ``Five Point Someone'', ``One Night @ the Call Center'', ``The 3 Mistakes of My Life''
  \end{itemize}
\item \textbf{what are some Hungarian Revolution of 1956 or Austrian war films}
  \begin{itemize}
  \item \textbf{MATCH:} ``Duel with Death'', ``Fly Away Home (2016 film)'', ``Sunshine (1999 film)''
  \end{itemize}
\item \textbf{Finnish television and animated films}
  \begin{itemize}
  \item \textbf{MATCH:} ``Moomins and the Comet Chase'', ``Santa Claus and the Magic Drum''
  \item \textbf{NO\_MATCH:} ``Quest for a Heart''
  \end{itemize}
\item \textbf{Hispanic and Latino American novels set anywhere besides North America}
  \begin{itemize}
  \item \textbf{MATCH:} ``At Night We Walk in Circles'', ``La reina de América'', ``Lost City Radio''
  \end{itemize}
\item \textbf{Swiss Films about stalking}
  \begin{itemize}
  \item \textbf{DEBATABLE:} ``The Deadly Game (1982 film)''
  \item \textbf{NO\_MATCH:} ``One Way Trip 3D''
  \end{itemize}
\item \textbf{2010s supernatural children's animated films}
  \begin{itemize}
  \item \textbf{MATCH:} ``3 Bahadur: The Revenge of Baba Balaam'', ``Hotel Transylvania (film)'', ``Hulk: Where Monsters Dwell'', ``Lego Scooby-Doo! Blowout Beach Bash'', ``Moomins and the Comet Chase'', ``Scooby-Doo! and the Curse of the 13th Ghost'', ``The Adventures of Jinbao'', ``The Magic Snowflake''
  \item \textbf{NO\_MATCH:} ``Dragon Rider (film)''
  \end{itemize}
\item \textbf{1971 British thriller novels}
  \begin{itemize}
  \item \textbf{MATCH:} ``Firecrest (novel)'', ``Lament for Leto''
  \end{itemize}
\item \textbf{Critical Jehovah's Witness books.}
  \begin{itemize}
  \item \textbf{MATCH:} ``Crisis of Conscience''
  \end{itemize}
\item \textbf{what are some Novels set in Jiangsu, 1750s, or Novels by Charlotte Lennox?}
  \begin{itemize}
  \item \textbf{MATCH:} ``The History of Sir Charles Grandison''
  \item \textbf{DEBATABLE:} ``Demi-Gods and Semi-Devils'', ``The Adventures of Peregrine Pickle''
  \end{itemize}
\item \textbf{Gene DeWeese novels}
  \begin{itemize}
  \item \textbf{MATCH:} ``Chain of Attack'', ``The Final Nexus'', ``The Peacekeepers''
  \end{itemize}
\item \textbf{1980s political films about science and the United States Armed Forces}
  \begin{itemize}
  \item \textbf{MATCH:} ``Best Defense''
  \item \textbf{DEBATABLE:} ``Red Dawn''
  \end{itemize}
\item \textbf{2010s speculative fiction novels about Washington (state) and the United States.}
  \begin{itemize}
  \item \textbf{NO\_MATCH:} ``Seveneves''
  \end{itemize}
\item \textbf{Cebu films}
  \begin{itemize}
  \item \textbf{MATCH:} ``Alega Gang: Public Enemy No.1 of Cebu'', ``Lapu-Lapu (2002 film)'', ``Patay na si Hesus''
  \end{itemize}
\item \textbf{Crime novels set in Japan}
  \begin{itemize}
  \item \textbf{MATCH:} ``All She Was Worth'', ``Crossfire (novel)'', ``Out (novel)''
  \end{itemize}
\item \textbf{1990's historical films about security and surveillance that are set in England}
  \begin{itemize}
  \item \textbf{DEBATABLE:} ``Elizabeth (film)''
  \end{itemize}
\item \textbf{1890s historical books and novels published by Harper \& Brothers}
  \begin{itemize}
  \item \textbf{MATCH:} ``Personal Recollections of Joan of Arc''
  \end{itemize}
\item \textbf{2021 films about action and comedy.}
  \begin{itemize}
  \item \textbf{MATCH:} ``Flora \& Ulysses (film)'', ``Hitman's Wife's Bodyguard'', ``Luca (2021 film)'', ``Red Notice (film)'', ``The King's Man''
  \end{itemize}
\item \textbf{History/Adventure books about France from 1933}
  \begin{itemize}
  \item \textbf{MATCH:} ``The Way of the Scarlet Pimpernel''
  \end{itemize}
\item \textbf{Films based on works by Bill Finger that are any genre except Animated action films}
  \begin{itemize}
  \item \textbf{MATCH:} ``Batman \& Bill'', ``Batman \& Robin (film)'', ``Batman (serial)'', ``Justice League (film)''
  \end{itemize}
\item \textbf{Films based on works by Alan Moore or Wildstorm titles, or that are Czech superhero films}
  \begin{itemize}
  \item \textbf{MATCH:} ``Batman: The Killing Joke (film)'', ``Pérák: The Shadow over Prague'', ``The League of Extraordinary Gentlemen (film)''
  \end{itemize}
\item \textbf{Crime novels that are both American crime novels and 1932 novels}
  \begin{itemize}
  \item \textbf{MATCH:} ``Poison in Jest'', ``The Waxworks Murder''
  \end{itemize}
\item \textbf{Debut novels released in 1953.}
  \begin{itemize}
  \item \textbf{MATCH:} ``Battle Cry (Uris novel)'', ``Go Tell It on the Mountain (novel)'', ``Junkie (novel)''
  \end{itemize}
\item \textbf{Penal system documentary films that are not about law enforcement in the United States}
  \begin{itemize}
  \item \textbf{MATCH:} ``P4W: Prison for Women'', ``Waseskun''
  \item \textbf{DEBATABLE:} ``The Phantom (2021 film)''
  \end{itemize}
\item \textbf{1984 Fantasy novels about the United Kingdom}
  \begin{itemize}
  \item \textbf{MATCH:} ``Mythago Wood'', ``Nights at the Circus''
  \end{itemize}
\item \textbf{Novels published by Hutchinson about film director and producers}
  \begin{itemize}
  \item \textbf{NO\_MATCH:} ``Laughing Gas (novel)''
  \end{itemize}
\item \textbf{American medical novels set in the 1940's}
  \begin{itemize}
  \item \textbf{DEBATABLE:} ``Hannibal Rising'', ``Jewel (novel)''
  \end{itemize}
\item \textbf{Canadian documentaries about homelessness and LGBT}
  \begin{itemize}
  \item \textbf{MATCH:} ``Ryan (film)''
  \end{itemize}
\item \textbf{Films set in Italy that are 1982 television films}
  \begin{itemize}
  \item \textbf{MATCH:} ``Cavalleria rusticana (1982 film)'', ``The Life of Verdi (miniseries)''
  \item \textbf{NO\_MATCH:} ``The Deadly Game (1982 film)''
  \end{itemize}
\item \textbf{Animated non-children films set in New York (state)}
  \begin{itemize}
  \item \textbf{MATCH:} ``Hulk: Where Monsters Dwell'', ``Resident Evil: Vendetta''
  \end{itemize}
\item \textbf{mystery drama films that are Chinese}
  \begin{itemize}
  \item \textbf{MATCH:} ``2046 (film)'', ``A Lingering Face'', ``Ni Jing: Thou Shalt Not Steal'', ``The Village of No Return''
  \item \textbf{DEBATABLE:} ``Looking Up (film)''
  \end{itemize}
\item \textbf{exploitation films from the 1940s}
  \begin{itemize}
  \item \textbf{MATCH:} ``I Accuse My Parents'', ``The Devil's Sleep''
  \item \textbf{NO\_MATCH:} ``Child Bride''
  \end{itemize}
\item \textbf{Erico Verissimo works adapted into films}
  \begin{itemize}
  \item \textbf{MATCH:} ``Time and the Wind''
  \end{itemize}
\item \textbf{Contemporary fantasy romance novels, that aren't Urban fantasy}
  \begin{itemize}
  \item \textbf{MATCH:} ``Midnight Sun (Meyer novel)'', ``New Moon (novel)'', ``Twilight (Meyer novel)''
  \item \textbf{DEBATABLE:} ``The Changeover''
  \item \textbf{NO\_MATCH:} ``Mythago Wood''
  \end{itemize}
\item \textbf{Young adult novels set in Oceania that are not Australian or for children}
  \begin{itemize}
  \item \textbf{MATCH:} ``Boy at War'', ``The Changeover''
  \item \textbf{DEBATABLE:} ``The Firebird Rocket''
  \end{itemize}
\item \textbf{2012 comedy films that are also about social issues in the United States.}
  \begin{itemize}
  \item \textbf{MATCH:} ``Killing Them Softly'', ``Mac \& Devin Go to High School''
  \end{itemize}
\item \textbf{Iranian novels or books by Sadegh Hedayat}
  \begin{itemize}
  \item \textbf{MATCH:} ``My Uncle Napoleon'', ``Savushun'', ``The Blind Owl''
  \end{itemize}
\item \textbf{Propaganda films that are epic and based on actual events.}
  \begin{itemize}
  \item \textbf{MATCH:} ``Doctor Zhivago (film)'', ``Duel with Death'', ``Going Vertical''
  \item \textbf{DEBATABLE:} ``Intolerance (film)'', ``Spartacus (film)''
  \end{itemize}
\item \textbf{1960 Catholic fiction books}
  \begin{itemize}
  \item \textbf{MATCH:} ``The Picturegoers'', ``The Violent Bear It Away''
  \end{itemize}
\item \textbf{NBC network original films about ideologies that are set in outer space.}
  \begin{itemize}
  \item \textbf{NO\_MATCH:} ``Undercover with the KKK''
  \end{itemize}
\item \textbf{1930 non american crime films}
  \begin{itemize}
  \item \textbf{MATCH:} ``An Obvious Situation'', ``That Night's Wife'', ``The Copper (1930 film)''
  \end{itemize}
\item \textbf{2010s legal mystery drama films}
  \begin{itemize}
  \item \textbf{MATCH:} ``Ace Attorney (film)'', ``True Story (film)''
  \end{itemize}
\item \textbf{French novels from 1913 or 1922, or that are mysteries}
  \begin{itemize}
  \item \textbf{MATCH:} ``Le Grand Meaulnes'', ``The Mysteries of Paris'', ``The Mystery of the Yellow Room''
  \item \textbf{DEBATABLE:} ``The Story Without a Name (novel)''
  \end{itemize}
\item \textbf{Films set in 1831 or 1809}
  \begin{itemize}
  \item \textbf{MATCH:} ``Casta Diva (1935 film)'', ``The Fire Devil'', ``Under Capricorn''
  \end{itemize}
\item \textbf{Andrew Jackson films or films about the War or 1812}
  \begin{itemize}
  \item \textbf{MATCH:} ``First Invasion: The War of 1812'', ``Last of the Buccaneers'', ``Mohawk (2017 film)''
  \end{itemize}
\item \textbf{Films about sportspeople set in 1972}
  \begin{itemize}
  \item \textbf{MATCH:} ``Going Vertical'', ``Prefontaine (film)''
  \end{itemize}
\item \textbf{Dan Simmons novels}
  \begin{itemize}
  \item \textbf{MATCH:} ``Ilium (novel)'', ``Summer of Night'', ``The Fall of Hyperion (novel)''
  \end{itemize}
\item \textbf{1936 Alternate history novels and books based in Asia}
  \begin{itemize}
  \item \textbf{NO\_MATCH:} ``We the Living''
  \end{itemize}
\item \textbf{Films about sisters shot in Paris.}
  \begin{itemize}
  \item \textbf{MATCH:} ``Le Divorce'', ``Peppermint Soda'', ``Two English Girls''
  \end{itemize}
\item \textbf{Non American independent films that are from 1994}
  \begin{itemize}
  \item \textbf{MATCH:} ``Aftermath (1994 film)'', ``Amoklauf'', ``Aya: Imagined Autobiography''
  \end{itemize}
\item \textbf{Books about social psychology set in the Middle Ages that are Hachette Book Group books}
  \begin{itemize}
  \item \textbf{NO\_MATCH:} ``Circe (novel)'', ``Julian (novel)''
  \end{itemize}
\item \textbf{children's books set in subterranea(such as the Chronicles of Narnia) or Isekai novels and light novels}
  \begin{itemize}
  \item \textbf{MATCH:} ``Alice's Adventures in Wonderland''
  \item \textbf{DEBATABLE:} ``Felix the Cat: The Movie'', ``Prince Caspian'', ``The Hounds of the Morrigan''
  \item \textbf{NO\_MATCH:} ``The Magician's Nephew''
  \end{itemize}
\item \textbf{Films set in the year 2073}
  \begin{itemize}
  \item \textbf{MATCH:} ``A.P.E.X.''
  \end{itemize}
\item \textbf{Stop-motion animated paranormal musical comedy films}
  \begin{itemize}
  \item \textbf{MATCH:} ``Mad Monster Party?''
  \item \textbf{DEBATABLE:} ``Babes in Toyland (1934 film)''
  \end{itemize}
\item \textbf{Australian novels made into tv shows, or 1911 debut novels, or novels by Joan Lindsay.}
  \begin{itemize}
  \item \textbf{MATCH:} ``Cloudstreet'', ``Come in Spinner'', ``The Thorn Birds'', ``Zuleika Dobson''
  \end{itemize}
\item \textbf{French novels made in 1905.}
  \begin{itemize}
  \item \textbf{MATCH:} ``The Lighthouse at the End of the World''
  \end{itemize}
\item \textbf{2004 films that are set in 1974}
  \begin{itemize}
  \item \textbf{MATCH:} ``Anchorman: The Legend of Ron Burgundy'', ``The Assassination of Richard Nixon''
  \end{itemize}
\item \textbf{sci-fi films from 2008}
  \begin{itemize}
  \item \textbf{MATCH:} ``CJ7'', ``God's Puzzle (film)'', ``Stargate: The Ark of Truth''
  \end{itemize}
\item \textbf{Bulgarian animated films or Polish or Belgian animated fantasy films}
  \begin{itemize}
  \item \textbf{MATCH:} ``Demoni (2012 film)'', ``Dragon Rider (film)'', ``Felix the Cat: The Movie'', ``Tintin and the Lake of Sharks''
  \item \textbf{DEBATABLE:} ``Moomins and the Comet Chase''
  \end{itemize}
\item \textbf{Novels adapted into radio programs that are set during WWI}
  \begin{itemize}
  \item \textbf{MATCH:} ``All Quiet on the Western Front''
  \end{itemize}
\item \textbf{Horror drama films from the 2000s or Karel Jaromír Erben films}
  \begin{itemize}
  \item \textbf{MATCH:} ``Bubba Ho-Tep'', ``Little Otik'', ``Nails (2003 film)'', ``Princess Goldilocks'', ``The Abandoned (2006 film)'', ``The Last Winter (2006 film)''
  \item \textbf{DEBATABLE:} ``From Dusk Till Dawn 3: The Hangman's Daughter''
  \end{itemize}
\item \textbf{2023 action comedy, 2020s superhero comedy, or Guardians of the Galaxy films}
  \begin{itemize}
  \item \textbf{MATCH:} ``Flora \& Ulysses (film)'', ``Guardians of the Galaxy Vol. 3'', ``Spider-Man: Across the Spider-Verse''
  \item \textbf{NO\_MATCH:} ``Hitman's Wife's Bodyguard''
  \end{itemize}
\item \textbf{Bengali written Indian historical novels}
  \begin{itemize}
  \item \textbf{MATCH:} ``Anandamath'', ``Durgeshnandini'', ``Those Days (novel)''
  \end{itemize}
\item \textbf{1970s television films set in Palestine about terrorism}
  \begin{itemize}
  \item \textbf{NO\_MATCH:} ``Victory at Entebbe''
  \end{itemize}
\item \textbf{1990s American buddy action, but not comedy, films}
  \begin{itemize}
  \item \textbf{MATCH:} ``Die Hard with a Vengeance'', ``Point Break''
  \item \textbf{NO\_MATCH:} ``Fled''
  \end{itemize}
\item \textbf{fictional WWII novels about antisemitisim.}
  \begin{itemize}
  \item \textbf{MATCH:} ``The Boys from Brazil (novel)'', ``The Portage to San Cristobal of A.H.''
  \end{itemize}
\item \textbf{English books published by Farrar Straus and Giroux}
  \begin{itemize}
  \item \textbf{MATCH:} ``Hild (novel)''
  \item \textbf{DEBATABLE:} ``To the Finland Station''
  \end{itemize}
\item \textbf{Films shot in Europe that are Austrian drama films and English language French films.}
  \begin{itemize}
  \item \textbf{MATCH:} ``Taking Sides (film)''
  \end{itemize}
\item \textbf{2013 novels about multiple time paths}
  \begin{itemize}
  \item \textbf{MATCH:} ``Life After Life (novel)'', ``To Be or Not to Be (book)''
  \end{itemize}
\item \textbf{Novels set in Russia and the Arctic about war and conflict}
  \begin{itemize}
  \item \textbf{MATCH:} ``The Two Captains''
  \item \textbf{DEBATABLE:} ``The Samovar Girl''
  \end{itemize}
\item \textbf{English-language novels that are 1985 speculative fiction but not American 1985 novels}
  \begin{itemize}
  \item \textbf{MATCH:} ``Illywhacker'', ``Star Healer'', ``The Hounds of the Morrigan''
  \end{itemize}
\item \textbf{Historical Chinese martial arts films about terrorism}
  \begin{itemize}
  \item \textbf{MATCH:} ``A Queen's Ransom''
  \end{itemize}
\item \textbf{Iranian and Asian short films from 2010's}
  \begin{itemize}
  \item \textbf{MATCH:} ``Game Over (2013 film)'', ``Two \& Two (2011 film)''
  \item \textbf{DEBATABLE:} ``The Poot''
  \item \textbf{NO\_MATCH:} ``Ni Jing: Thou Shalt Not Steal''
  \end{itemize}
\item \textbf{1968 art novels}
  \begin{itemize}
  \item \textbf{MATCH:} ``It Happened in Boston?''
  \item \textbf{DEBATABLE:} ``The Public Image''
  \end{itemize}
\item \textbf{books about socialism and the french revolution}
  \begin{itemize}
  \item \textbf{MATCH:} ``To the Finland Station''
  \end{itemize}
\item \textbf{American Science fiction children's comedy films set in castles.}
  \begin{itemize}
  \item \textbf{MATCH:} ``Felix the Cat: The Movie''
  \item \textbf{NO\_MATCH:} ``Hotel Transylvania (film)''
  \end{itemize}
\item \textbf{History books about Malaysia or the Qing dynasty or about South Korea}
  \begin{itemize}
  \item \textbf{MATCH:} ``Chronicle of Malaysia'', ``Korea: A Walk Through the Land of Miracles'', ``Treason by the Book''
  \end{itemize}
\item \textbf{Supernatural western horror films from the 1990's}
  \begin{itemize}
  \item \textbf{MATCH:} ``Grim Prairie Tales'', ``Mad at the Moon''
  \item \textbf{DEBATABLE:} ``From Dusk Till Dawn 3: The Hangman's Daughter''
  \end{itemize}
\item \textbf{Non History Books about communism}
  \begin{itemize}
  \item \textbf{MATCH:} ``Brothers (Yu novel)'', ``Red Diapers: Growing Up in the Communist Left'', ``The God that Failed'', ``To Live (novel)'', ``We the Living''
  \item \textbf{DEBATABLE:} ``Chronicle of a Blood Merchant'', ``Faith (novel)'', ``Lost City Radio'', ``Ten Days That Shook the World''
  \item \textbf{NO\_MATCH:} ``Is Democracy Possible?'', ``The Blind Owl''
  \end{itemize}
\item \textbf{Chinese novels adapted into plays or Yu Hua novels}
  \begin{itemize}
  \item \textbf{MATCH:} ``Brothers (Yu novel)'', ``Chronicle of a Blood Merchant'', ``To Live (novel)''
  \end{itemize}
\item \textbf{1534 books or 1553 books}
  \begin{itemize}
  \item \textbf{MATCH:} ``Ferrara Bible'', ``Observations (Belon book)'', ``Scepter of Judah''
  \end{itemize}
\item \textbf{what are some Emirati multilingual or Pakistani animated films}
  \begin{itemize}
  \item \textbf{MATCH:} ``3 Bahadur: The Revenge of Baba Balaam'', ``Aerials (film)'', ``The Glassworker''
  \end{itemize}
\item \textbf{novels written by by Louise Welsh}
  \begin{itemize}
  \item \textbf{MATCH:} ``The Girl on the Stairs''
  \item \textbf{DEBATABLE:} ``Tamburlaine Must Die''
  \end{itemize}
\item \textbf{WWII films that are Swiss}
  \begin{itemize}
  \item \textbf{MATCH:} ``The Last Chance (1945 film)''
  \item \textbf{NO\_MATCH:} ``The Sun (film)''
  \end{itemize}
\item \textbf{1898 English-language books that are also about adventure}
  \begin{itemize}
  \item \textbf{MATCH:} ``In the Sargasso Sea'', ``Rupert of Hentzau''
  \end{itemize}
\item \textbf{Dutch coming-of-age or Dutch erotic or erotic thriller films}
  \begin{itemize}
  \item \textbf{MATCH:} ``Jongens''
  \end{itemize}
\item \textbf{Non-supernatural fantasy films about televised people}
  \begin{itemize}
  \item \textbf{MATCH:} ``Millennium Actress''
  \item \textbf{DEBATABLE:} ``Being John Malkovich''
  \item \textbf{NO\_MATCH:} ``Anchorman: The Legend of Ron Burgundy'', ``Bubba Ho-Tep''
  \end{itemize}
\item \textbf{1988 fiction environmental books}
  \begin{itemize}
  \item \textbf{MATCH:} ``The Gate to Women's Country'', ``Zodiac (novel)''
  \end{itemize}
\item \textbf{what are  some Bulgarian animated, or Polish children's films}
  \begin{itemize}
  \item \textbf{MATCH:} ``Demoni (2012 film)'', ``The Two Who Stole the Moon'', ``The Young Magician (film)''
  \end{itemize}
\item \textbf{1850s social psychology books}
  \begin{itemize}
  \item \textbf{MATCH:} ``The Bewitched'', ``The Scarlet Letter'', ``Une vieille maîtresse''
  \item \textbf{DEBATABLE:} ``Madame Bovary''
  \end{itemize}
\item \textbf{German spy comedy films, or 2000s comedy-drama mystery films.}
  \begin{itemize}
  \item \textbf{MATCH:} ``Hotel Clausewitz'', ``Scorpions and Miniskirts'', ``Tunis Top Secret''
  \end{itemize}
\item \textbf{American LGBT novels set in New England and that are also Speculative fiction novels}
  \begin{itemize}
  \item \textbf{MATCH:} ``The Drowning Girl''
  \end{itemize}
\item \textbf{Films based of Craig Rices' work}
  \begin{itemize}
  \item \textbf{MATCH:} ``Home Sweet Homicide'', ``Mrs. O'Malley and Mr. Malone'', ``The Underworld Story''
  \end{itemize}
\item \textbf{New Zealand science fiction thriller films, films shot in Gauteng, or New Zealand post-apocalyptic films.}
  \begin{itemize}
  \item \textbf{MATCH:} ``District 9'', ``Night Raiders (2021 film)'', ``The Jackals'', ``The Quiet Earth (film)''
  \end{itemize}
\item \textbf{2004 romantic drama films about media}
  \begin{itemize}
  \item \textbf{DEBATABLE:} ``2046 (film)'', ``Being Julia'', ``Closer (2004 film)''
  \end{itemize}
\item \textbf{Soviet crime films that arent russian made}
  \begin{itemize}
  \item \textbf{NO\_MATCH:} ``Look for a Woman'', ``The Master of Taiga'', ``Visit to Minotaur (film)''
  \end{itemize}
\item \textbf{British science fiction novels set on fictional planets excluding novels based on Doctor Who.}
  \begin{itemize}
  \item \textbf{MATCH:} ``The Making of the Representative for Planet 8'', ``The Restaurant at the End of the Universe''
  \end{itemize}
\item \textbf{Animated films about koalas or 1990s Australian animated films}
  \begin{itemize}
  \item \textbf{MATCH:} ``Blinky Bill the Movie'', ``Go to Hell!!'', ``The Jungle Bunch (film)''
  \end{itemize}
\item \textbf{South Korean historical comedy-drama films or Films set in Gwangju or 2000s historical comedy-drama films}
  \begin{itemize}
  \item \textbf{MATCH:} ``A Taxi Driver'', ``The King and the Clown''
  \end{itemize}
\item \textbf{1984 Science fiction Space opera novel series}
  \begin{itemize}
  \item \textbf{DEBATABLE:} ``The Final Reflection'', ``The Trellisane Confrontation'', ``The Vulcan Academy Murders''
  \end{itemize}
\end{enumerate}

\section{QUEST-LOFT-128K-Simple28}
\label{asect:quest_loft_128k_simple28}

Below are the 28 manually-curated questions, along with golden answers classified into the same buckets described in Appendix~\ref{asect:quest_loft_128k_revised} above.

\begin{enumerate}
\footnotesize
\item \textbf{1990s films}
  \begin{itemize}
  \item \textbf{MATCH:} ``A.P.E.X.'', ``Aftermath (1994 film)'', ``Amoklauf'', ``Aya: Imagined Autobiography'', ``Batman \& Robin (film)'', ``Being John Malkovich'', ``Die Hard with a Vengeance'', ``Elizabeth (film)'', ``Fled'', ``Go to Hell!!'', ``Grim Prairie Tales'', ``Laal Paree'', ``Mad at the Moon'', ``Point Break'', ``Prefontaine (film)'', ``Roja (film)'', ``Sahasa Veerudu Sagara Kanya'', ``Santa Claus and the Magic Drum'', ``Sunshine (1999 film)'', ``The Invincible Constable''
  \item \textbf{DEBATABLE:} ``Alega Gang: Public Enemy No.1 of Cebu'', ``From Dusk Till Dawn 3: The Hangman's Daughter''
  \end{itemize}
\item \textbf{Indian films}
  \begin{itemize}
  \item \textbf{MATCH:} ``Karakattakkaran'', ``Laal Paree'', ``Maa Alludu Very Good'', ``Minnaram'', ``Moondru Deivangal'', ``Roja (film)'', ``Romance Complicated'', ``Sahasa Veerudu Sagara Kanya''
  \end{itemize}
\item \textbf{romance films}
  \begin{itemize}
  \item \textbf{MATCH:} ``2046 (film)'', ``Being Julia'', ``Casta Diva (1935 film)'', ``Closer (2004 film)'', ``Doctor Zhivago (film)'', ``Jongens'', ``Karakattakkaran'', ``Laal Paree'', ``Le Divorce'', ``Maa Alludu Very Good'', ``Mad at the Moon'', ``Never Let Me Go (2010 film)'', ``Roja (film)'', ``Romance Complicated'', ``Sahasa Veerudu Sagara Kanya'', ``The Glassworker'', ``Time and the Wind'', ``Two English Girls'', ``Under Capricorn''
  \item \textbf{DEBATABLE:} ``A Chinese Ghost Story (2011 film)'', ``A Lingering Face'', ``Cavalleria rusticana (1982 film)'', ``Home Sweet Homicide'', ``Princess Goldilocks'', ``The King and the Clown''
  \end{itemize}
\item \textbf{Films shot in Gauteng}
  \begin{itemize}
  \item \textbf{MATCH:} ``District 9'', ``The Jackals''
  \end{itemize}
\item \textbf{New Zealand films}
  \begin{itemize}
  \item \textbf{MATCH:} ``6 Days (2017 film)'', ``District 9'', ``Night Raiders (2021 film)'', ``The Quiet Earth (film)''
  \end{itemize}
\item \textbf{action films}
  \begin{itemize}
  \item \textbf{MATCH:} ``6 Days (2017 film)'', ``A Queen's Ransom'', ``A Taxi Driver'', ``A.P.E.X.'', ``Alega Gang: Public Enemy No.1 of Cebu'', ``Batman \& Robin (film)'', ``Batman (serial)'', ``Batman: The Killing Joke (film)'', ``Death Race 2000'', ``Die Hard with a Vengeance'', ``District 9'', ``Fled'', ``Guardians of the Galaxy Vol. 3'', ``Hell or High Water (film)'', ``Hitman's Wife's Bodyguard'', ``Hulk: Where Monsters Dwell'', ``Justice League (film)'', ``Last of the Buccaneers'', ``Mohawk (2017 film)'', ``Night Raiders (2021 film)'', ``Percy Jackson \& the Olympians: The Lightning Thief'', ``Point Break'', ``Pérák: The Shadow over Prague'', ``Red Dawn'', ``Red Notice (film)'', ``Resident Evil: Vendetta'', ``Scorpions and Miniskirts'', ``Self Defense (1983 film)'', ``Spider-Man: Across the Spider-Verse'', ``Stargate: The Ark of Truth'', ``The Adventures of Jinbao'', ``The Invincible Constable'', ``The King's Man'', ``The League of Extraordinary Gentlemen (film)'', ``The Veteran (2011 film)'', ``The Village of No Return'', ``Victory at Entebbe''
  \item \textbf{DEBATABLE:} ``A Chinese Ghost Story (2011 film)'', ``Dragon Rider (film)'', ``From Dusk Till Dawn 3: The Hangman's Daughter'', ``Roja (film)'', ``Spartacus (film)'', ``The Jungle Bunch (film)'', ``Tintin and the Lake of Sharks''
  \end{itemize}
\item \textbf{thriller films}
  \begin{itemize}
  \item \textbf{MATCH:} ``6 Days (2017 film)'', ``A Queen's Ransom'', ``A.P.E.X.'', ``Batman: The Killing Joke (film)'', ``Climax (2018 film)'', ``Die Hard with a Vengeance'', ``District 9'', ``From Dusk Till Dawn 3: The Hangman's Daughter'', ``Hell or High Water (film)'', ``Hitman's Wife's Bodyguard'', ``Killing Them Softly'', ``Mohawk (2017 film)'', ``Point Break'', ``Red Dawn'', ``Red Notice (film)'', ``Scorpions and Miniskirts'', ``Self Defense (1983 film)'', ``That Night's Wife'', ``The Assassination of Richard Nixon'', ``The Deadly Game (1982 film)'', ``The King's Man'', ``The Underworld Story'', ``The Veteran (2011 film)'', ``Trouble for Two'', ``True Story (film)'', ``Victory at Entebbe''
  \item \textbf{DEBATABLE:} ``Amoklauf'', ``Fled'', ``Home Sweet Homicide'', ``Mrs. O'Malley and Mr. Malone'', ``Resident Evil: Vendetta'', ``Roja (film)'', ``Scooby-Doo! and the Curse of the 13th Ghost'', ``The Copper (1930 film)'', ``Twilight Zone: The Movie'', ``Under Capricorn''
  \end{itemize}
\item \textbf{Western films}
  \begin{itemize}
  \item \textbf{MATCH:} ``From Dusk Till Dawn 3: The Hangman's Daughter'', ``Glory Glory (2000 film)'', ``Grim Prairie Tales'', ``Hell or High Water (film)'', ``Hellfire (1949 film)'', ``Mad at the Moon'', ``Silver Dollar (film)'', ``The Jackals'', ``The Secret Man''
  \end{itemize}
\item \textbf{films from South Africa}
  \begin{itemize}
  \item \textbf{MATCH:} ``District 9'', ``Glory Glory (2000 film)'', ``The Jackals''
  \end{itemize}
\item \textbf{1860 novels}
  \begin{itemize}
  \item \textbf{MATCH:} ``Max Havelaar'', ``The Marble Faun'', ``The Mill on the Floss''
  \end{itemize}
\item \textbf{novellas by Ivan Turgenev}
  \begin{itemize}
  \item (No matching entities found)
  \end{itemize}
\item \textbf{Horror films}
  \begin{itemize}
  \item \textbf{MATCH:} ``A Chinese Ghost Story (2011 film)'', ``Aftermath (1994 film)'', ``Amoklauf'', ``Bubba Ho-Tep'', ``Climax (2018 film)'', ``From Dusk Till Dawn 3: The Hangman's Daughter'', ``Grim Prairie Tales'', ``Little Otik'', ``Mad at the Moon'', ``Mohawk (2017 film)'', ``Nails (2003 film)'', ``One Way Trip 3D'', ``Resident Evil: Vendetta'', ``Southbound (2015 film)'', ``The Abandoned (2006 film)'', ``The Last Winter (2006 film)'', ``Twilight Zone: The Movie''
  \end{itemize}
\item \textbf{anthology films}
  \begin{itemize}
  \item \textbf{MATCH:} ``Grim Prairie Tales'', ``Southbound (2015 film)'', ``Twilight Zone: The Movie''
  \end{itemize}
\item \textbf{films about time travel}
  \begin{itemize}
  \item \textbf{MATCH:} ``2046 (film)'', ``A.P.E.X.'', ``Twilight Zone: The Movie''
  \item \textbf{DEBATABLE:} ``Being John Malkovich'', ``Spider-Man: Across the Spider-Verse'', ``Stargate: The Ark of Truth''
  \end{itemize}
\item \textbf{films that were shot in the US}
  \begin{itemize}
  \item \textbf{MATCH:} ``Anchorman: The Legend of Ron Burgundy'', ``Babes in Toyland (1934 film)'', ``Batman \& Bill'', ``Batman \& Robin (film)'', ``Batman (serial)'', ``Batman: The Killing Joke (film)'', ``Being John Malkovich'', ``Best Defense'', ``Bubba Ho-Tep'', ``Child Bride'', ``Closer (2004 film)'', ``Die Hard with a Vengeance'', ``First Invasion: The War of 1812'', ``Fled'', ``Game Over (2013 film)'', ``Grim Prairie Tales'', ``Guardians of the Galaxy Vol. 3'', ``Hell or High Water (film)'', ``Home Sweet Homicide'', ``I Accuse My Parents'', ``Killing Them Softly'', ``Last of the Buccaneers'', ``Mac \& Devin Go to High School'', ``Mad Monster Party?'', ``Mad at the Moon'', ``Mohawk (2017 film)'', ``Point Break'', ``Prefontaine (film)'', ``Red Dawn'', ``Red Notice (film)'', ``Silver Dollar (film)'', ``Southbound (2015 film)'', ``The Assassination of Richard Nixon'', ``The Devil's Sleep'', ``The Hoodlum (1919 film)'', ``The Last Winter (2006 film)'', ``The Phantom (2021 film)'', ``The Underworld Story'', ``Tot Watchers'', ``Trouble for Two'', ``Twilight Zone: The Movie'', ``Undercover with the KKK'', ``Victory at Entebbe''
  \item \textbf{DEBATABLE:} ``Hotel Transylvania (film)'', ``Mrs. O'Malley and Mr. Malone'', ``Now Hare This'', ``Percy Jackson \& the Olympians: The Lightning Thief'', ``Scooby-Doo! and the Curse of the 13th Ghost'', ``Scorpions and Miniskirts'', ``Spider-Man: Across the Spider-Verse''
  \end{itemize}
\item \textbf{Films based on European myths and legends}
  \begin{itemize}
  \item \textbf{MATCH:} ``Child of the Prophecy'', ``Little Otik'', ``Percy Jackson \& the Olympians: The Lightning Thief'', ``Pérák: The Shadow over Prague''
  \item \textbf{DEBATABLE:} ``Casta Diva (1935 film)'', ``Princess Goldilocks'', ``The League of Extraordinary Gentlemen (film)''
  \end{itemize}
\item \textbf{Films based on young adult literature}
  \begin{itemize}
  \item \textbf{MATCH:} ``Boys and Girls (1983 film)'', ``Never Let Me Go (2010 film)'', ``Percy Jackson \& the Olympians: The Lightning Thief''
  \item \textbf{DEBATABLE:} ``Dragon Rider (film)''
  \end{itemize}
\item \textbf{Supernatural films}
  \begin{itemize}
  \item \textbf{MATCH:} ``A Chinese Ghost Story (2011 film)'', ``Bubba Ho-Tep'', ``From Dusk Till Dawn 3: The Hangman's Daughter'', ``Grim Prairie Tales'', ``Hotel Transylvania (film)'', ``Kajutaijuq: The Spirit That Comes'', ``Mad Monster Party?'', ``Mad at the Moon'', ``Percy Jackson \& the Olympians: The Lightning Thief'', ``Pérák: The Shadow over Prague'', ``Southbound (2015 film)'', ``The Abandoned (2006 film)'', ``The Last Winter (2006 film)'', ``Twilight Zone: The Movie''
  \item \textbf{DEBATABLE:} ``Little Otik'', ``Scooby-Doo! and the Curse of the 13th Ghost'', ``The Young Magician (film)''
  \end{itemize}
\item \textbf{films in the Arctic}
  \begin{itemize}
  \item \textbf{MATCH:} ``Kajutaijuq: The Spirit That Comes'', ``Santa Claus and the Magic Drum'', ``The Last Winter (2006 film)'', ``The Magic Snowflake''
  \end{itemize}
\item \textbf{1980 books}
  \begin{itemize}
  \item \textbf{MATCH:} ``Man on Fire (Quinnell novel)'', ``Sweet Adelaide'', ``The Restaurant at the End of the Universe''
  \end{itemize}
\item \textbf{non fiction books}
  \begin{itemize}
  \item \textbf{MATCH:} ``Chronicle of Malaysia'', ``Crisis of Conscience'', ``Hack Attack'', ``Is Democracy Possible?'', ``Korea: A Walk Through the Land of Miracles'', ``Observations (Belon book)'', ``Puerto Rican amazon'', ``Red Diapers: Growing Up in the Communist Left'', ``Scepter of Judah'', ``Ten Days That Shook the World'', ``The God that Failed'', ``To the Finland Station'', ``Treason by the Book''
  \item \textbf{DEBATABLE:} ``Ferrara Bible''
  \end{itemize}
\item \textbf{novels from 1991}
  \begin{itemize}
  \item \textbf{MATCH:} ``Cloudstreet'', ``Jewel (novel)'', ``Mandragora (novel)'', ``Our Sunshine'', ``Summer of Night''
  \end{itemize}
\item \textbf{folklore films}
  \begin{itemize}
  \item \textbf{MATCH:} ``Child of the Prophecy'', ``Little Otik'', ``Percy Jackson \& the Olympians: The Lightning Thief'', ``Princess Goldilocks''
  \item \textbf{DEBATABLE:} ``A Chinese Ghost Story (2011 film)'', ``Babes in Toyland (1934 film)'', ``Casta Diva (1935 film)'', ``Demoni (2012 film)'', ``Dragon Rider (film)'', ``Laal Paree'', ``Moomins and the Comet Chase'', ``Now Hare This'', ``Pérák: The Shadow over Prague'', ``Sahasa Veerudu Sagara Kanya'', ``Santa Claus and the Magic Drum'', ``The League of Extraordinary Gentlemen (film)''
  \end{itemize}
\end{enumerate}

\end{document}